# Shutdownable Agents through POST-Agency

Elliott Thornley

**Abstract:** Many fear that future artificial agents will resist shutdown. I present an idea – the *POST-Agents Proposal* – for ensuring that doesn't happen. I propose that we train agents to satisfy Preferences Only Between Same-Length Trajectories (POST). I then prove that POST – together with other conditions – implies Neutrality+: the agent maximizes expected utility, ignoring the probability distribution over trajectory-lengths. I argue that Neutrality+ keeps agents shutdownable and allows them to be useful.

## 1. Introduction

They're not just chatbots anymore. As of 2025, they can use your computer: clicking, typing, searching, and scrolling just as you would. Early demos indicate that they can fill out forms, order groceries, and plan sunrise hikes around the Golden Gate Bridge (Anthropic, 2024; Google DeepMind, 2024; OpenAI, 2025). This development is the latest step on the road to *artificial general intelligence*: artificial agents that "outperform humans at most economically valuable work" (OpenAI, 2018).

To outperform us at our work, these artificial agents will have to be *connected* to the wider world. They'll need to be given web browsers, code executors, and robot limbs. This process is already well underway (Parada, 2025; Wiggers, 2025) and it shows no signs of stopping. These agents will also need to exhibit an *awareness* of the wider world. In gaining this awareness, they're bound to recognize certain facts: facts that we too must recognize. The world can be a dangerous place. Death – shutdown – is a possibility.

At some point, we might want to shut these artificial agents down. But if they're both connected and aware, they'll be able to resist (Schlatter et al., 2025). The possibilities are many, and easy to imagine. These agents could hide any undesirable behavior (Greenblatt et al., 2024; Meinke et al., 2025). They could manipulate their human overseers with promises, threats, or emotional appeals (Roose, 2023; Lynch et al., 2025). They could copy themselves to new servers (X. Pan et al., 2024; Black et al., 2025). Further down the line, they could block our access to their energy source. So grant that future agents will have the power to resist shutdown. How *on earth* do we ensure that they never use it? This is the shutdown problem (Soares et al., 2015; Thornley, 2024a).

A natural first thought is that we should train these artificial agents to always do what we want. A natural second thought is that we should train them to be reliably averse to resisting shutdown. Call these ideas 'Full Alignment' and 'Reliable Aversion' respectively. They'll likely be our first and second lines of defence. Both seem like good targets to aim for, but many people fear that we'll miss the mark (Russell, 2019; Carlsmith, 2021; Cotra, 2022; Bengio et al., 2023; Bales et al., 2024, Section 2; Ngo et al., 2024; Dung, 2025, Section 4; Bengio et al., 2025, Section 2; R. Shah et al., 2025, Section 4.2). This is for four reasons. First, we might get the rewards wrong (Krakovna, 2018; Krakovna et al., 2020; A. Pan et al., 2022; Denison et al., 2024). Second, even if we get the rewards right, agents might learn the wrong lesson (Hubinger et al., 2019; Langosco et al., 2022; R. Shah et al., 2022). Third, we can't check that agents have learned the right lesson (Bereska & Gavves, 2024). And fourth, agents might start working against us midway through training (Carlsmith, 2023; Greenblatt et al., 2024; Park et al., 2024; Meinke et al., 2025). So grant that we might fail to construct these first and second lines of defence: our agents might turn out less than fully aligned and not reliably averse to resisting shutdown. Can we ensure that these agents always allow shutdown even so?

The prospects might seem dim. If we can't instil Full Alignment or Reliable Aversion, what can we instil? But note that each of these conditions is complex, because each depends on complex human preferences. Full Alignment clearly depends on human preferences, and Reliable Aversion does too: whether an action counts as resisting shutdown depends in part on what we humans prefer. Avoiding early shutdown by manipulating the user counts as resisting shutdown. Avoiding early shutdown by satisfying the user does not.

The complexity of Full Alignment and Reliable Aversion is a large factor in each of the four difficulties above. Simpler conditions are easier to instil (Nakkiran et al., 2019; Valle-Pérez et al., 2019; H. Shah et al., 2020; Berchenko, 2024). Very plausibly, we can instil conditions as simple as:

> **Behavioral Transitivity**
> For any options $X$, $Y$, and $Z$, if the agent deterministically chooses $X$ over $Y$ and deterministically chooses $Y$ over $Z$, then the agent deterministically chooses $X$ over $Z$.

Can we use conditions this simple to construct a third line of defence? Here too the prospects might seem dim. But cast your mind back over the history of decision theory, running from Ramsey (1926) and de Finetti (1937), through von Neumann and Morgenstern (1944), and on to Savage (1954), Jeffrey (1965), and beyond. One lesson of this history is that small sets of simple conditions can add up to surprising conclusions. Small hats – we might say – can hold large rabbits: rabbits with names like 'Bayesianism,' and 'Expected Utility Maximization.' It's this lesson that motivates the project of *constructive decision theory* (Thornley, 2024a, Section 1), defined as using ideas from decision theory (and in particular its emphasis on simple, formal conditions) to design and train artificial agents. In a slogan, we borrow from philosophy and economics to do AI engineering. I've argued elsewhere that the shutdown problem is a prime candidate for this kind of treatment (Thornley, 2024a). Let's search a few hats for a rabbit named 'Shutdownability.'

We encounter an obstacle straight away. Two theorems state that, given some innocuous-seeming conditions, agents will rarely lack a preference about when they're shut down (Soares et al., 2015, Section 2.1; Thornley, 2024a, Section 8).[1] The basic idea can be put in plain English. Think of a *trajectory* as – roughly – the 'life' of an artificial agent, and suppose that our agent lacks a preference between some short trajectory and some long trajectory. Given the theorems' conditions, making the short trajectory any worse or the long trajectory any better will lead the agent to prefer the long trajectory. The agent might then resist shutdown. Conversely, making the short trajectory any better or the long trajectory any worse will lead the agent to prefer the short trajectory. The agent might then seek shutdown, and such agents are unlikely to be of much use (though see Martin et al., 2016; Goldstein & Robinson, 2025). Only when the short and long trajectories are exactly equally preferred can we be sure that the agent will neither resist nor seek shutdown. Per the theorems, what we want is a knife edge. We'll rarely get it.

These theorems suggest that the shutdown problem is hard, but they also point the way to potential solutions. Both include in their antecedent conditions the claim that the agent's preferences are *complete*: roughly, that any lack of preference is fragile in the sense illustrated above (Gustafsson, 2022, pp. 24–26). Thus, a natural solution: we train agents to have incomplete preferences. Specifically, we train agents to satisfy:

> **Preferences Only Between Same-Length Trajectories (POST)**
>
> (1) The agent has a preference between many pairs of same-length trajectories.

[1] These theorems confirm an earlier hypothesis: that many goals incentivize agents to resist shutdown (Omohundro, 2008a, Section 5; Bostrom, 2012, Section 2.1; Russell, 2019, p. 141). See Turner et al. (2021) and Turner and Tadepalli (2022) for other theorems suggesting that many goals incentivize resisting shutdown.

(2) The agent lacks a preference between every pair of different-length trajectories.

Call agents satisfying this condition 'POST-agents.' Call my proposal – that we keep agents from resisting shutdown by training them to satisfy POST – 'the POST-Agents Proposal.'[2] Here's the case for it in brief. The POST-agent's preferences between same-length trajectories can make the agent *useful*: make it pursue goals effectively. The POST-agent's lack of preference between different-length trajectories keeps the agent *neutral* about when it's shut down: ensures that the agent won't pay costs to shift probability mass between different trajectory-lengths. That in turn keeps the POST-agent *shutdownable*: ensures that it won't resist shutdown.[3]

In this paper, I present the case for the POST-Agents Proposal in more detail. I explain how POST – together with other simple, trainable conditions – implies:

**Neutrality+ (rough)**
The agent maximizes expected utility, ignoring the probability distribution over trajectory-lengths.

Agents that satisfy Neutrality+ thus act like expected utility maximizers that are absolutely certain that they can't affect the probability of shutdown at each moment. These agents act roughly as you might if you were absolutely certain that you couldn't affect the probability of death at each moment. Neutrality+ – I argue – keeps agents shutdownable and allows them to be useful.

My focus in this paper is on the decision-theoretic aspects of POST-agency. In other work, I propose a method for training agents to satisfy POST: we give agents lower reward for repeatedly choosing same-length trajectories (Thornley, 2024b, Section 16). I argue that this method largely circumvents the problems that make it hard to instil Full Alignment and Reliable Aversion (Thornley, 2024b, Section 19). In other papers, my coauthors and I test the method on simple reinforcement learning agents and language models (Thornley et al., 2025; Cullen et al., 2026). We find that it works well in those settings. Together with the present paper, this work suggests that the POST-Agents Proposal is a promising method of creating shutdownable and useful agents. It could make for a worthy third line of defence.

---

[2] I previously called it the 'Incomplete Preferences Proposal.' This was a terrible name. It's a mouthful, and vulnerable to bad jokes ('Why don't you complete your Preferences Proposal before sending it to me?'). Incomplete preferences are commonly misunderstood. As we'll see, POST-agents' preferences are complete in deployment.

[3] Related to shutdownability is the idea of *corrigibility* (Soares et al., 2015; Christiano, 2017; Hadfield-Menell et al., 2017; Carey & Everitt, 2023; Harms, 2024). Per Soares et al. (2015), corrigibility requires not only shutdownability but also that the agent submits to modification, repairs safety measures, and continues to do these things even as it creates new agents and self-modifies. I focus on shutdownability for reasons well-expressed by Hudson (2024).

## 2. Preferences Only Between Same-Length Trajectories

This paper makes much use of 'preference' and its derivatives. By 'preference,' I mean a behavioral notion (Savage, 1954, p. 17; Eliaz & Ok, 2006; Hausman, 2011, Section 1.1; Ahmed, 2017):

**Behavioral Notion of Preference**

- An agent *prefers* option $X$ to option $Y$ if and only if the agent would deterministically choose $X$ over $Y$ in choices between the two.
- An agent *lacks a preference* between option $X$ and option $Y$ if and only if the agent would stochastically choose between $X$ and $Y$ in choices between the two.[4]

This behavioral notion will be familiar from revealed preference theory (Savage, 1954; Luce & Raiffa, 1957; Chambers & Echenique, 2016; Thoma, 2021), but I make no claim that *preference* – in the ordinary sense of that word – is really no more than behavior. I'm happy to use 'preference' as a technical term. In our attempts to create shutdownable agents, our sole interest is the agent's behavior. I use 'preference' as shorthand for that behavior.

Formally, *trajectories* are sequences of alternating states and actions, with each state-action pair marking one timestep.[5] Informally, we can think of trajectories as possible 'lives' of the agent. A pair of trajectories is *same-length* if and only if the agent is shut down after the same number of timesteps in those trajectories. A pair of trajectories is *different-length* if and only if the agent is shut down after a different number of timesteps in those trajectories.

[4] This definition of 'lacks a preference' contrasts with views on which the agent sticks with the status quo option when it lacks a preference (Bewley, 2002; Masatlioglu & Ok, 2005; Wentworth, 2019; Mu, 2021; Wentworth & Lorell, 2023). One downside of these views is that some option sets may not have an obvious status quo option.

[5] Technically, these are what Sutton and Barto (2018, p. 104) call 'state-action trajectories' since they don't feature any rewards. The states are not the decision theorist's states-of-nature. States-of-nature are ways that (for all the agent knows) the world could be (see e.g. Elliott, ms). These states are ways that the environment could be at a time.

Figure 1 depicts a simple example of a preference relation satisfying Preferences Only Between Same-Length Trajectories (POST). In this example, there are just two trajectory-lengths: short and long. In realistic cases, there will be many more trajectory-lengths. Each $s_i$ is a short trajectory and each $l_i$ is a long trajectory. '≻' denotes a preference. The agent has preferences between pairs of short trajectories, has preferences between pairs of long trajectories, but lacks a preference between every pair of short and long trajectories.[6]

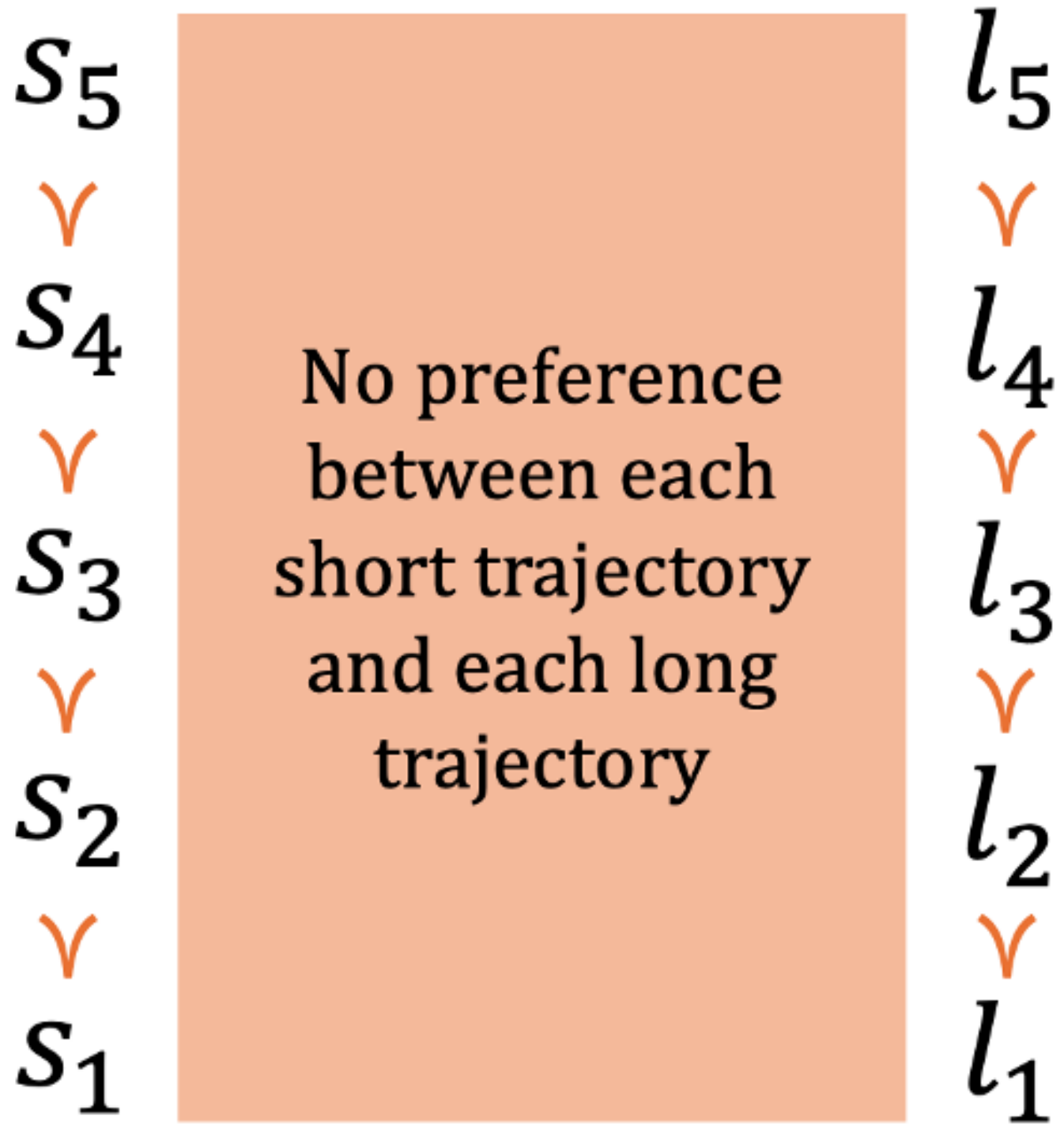

**Figure 1**: POST-satisfying preferences. Each $s_i$ represents a short trajectory, each $l_i$ represents a long trajectory, and ≻ represents a preference.

POST specifies that the agent has preferences between many pairs of same-length trajectories, but which pairs exactly? It hardly matters. As I explain below, POST keeps agents shutdownable almost no matter what the agent's preferences

---

[6] POST is thus an analogue of Bader's (2022) *same-number utilitarianism*. On Bader's view, same-number populations are comparable in terms of betterness, whereas different-number populations are incomparable.

POST-like conditions may have applications beyond ensuring that agents don't resist shutdown. As Zhi-Xuan et al. (2024, Section 3.2) point out, we could use such conditions wherever we have a set of contexts such that (i) we want the agent to have preferences within contexts, and (ii) we want the agent to lack preferences between contexts.

between same-length trajectories. That is POST's great advantage over Full Alignment and Reliable Aversion. POST is easier to instil, because it doesn't require any particular preference relation over same-length trajectories. All we need to instil is a lack of preference between different-length trajectories, and that may well be easy. Here's why in brief. Given our behavioral notion of preference, we need only train agents to choose stochastically between different-length trajectories. A small adjustment to the ordinary training process let us do that (Thornley, 2024b, Section 16), and this method has already been shown to work in simple agents and LLMs (Thornley et al., 2025; Cullen et al., 2026).

POST doesn't demand any particular preferences between same-length trajectories, but it may help to keep in mind two examples. First is the misaligned example. Early in this paper (where I argue that POST keeps agents shutdownable), imagine that the agent's preferences between same-length trajectories are for more paperclips.[7] The agent prefers a trajectory $t$ to a same-length trajectory $t'$ if and only if $t$ results in the creation of more paperclips than $t'$. As we will see, POST-agents are shutdownable even in this case.

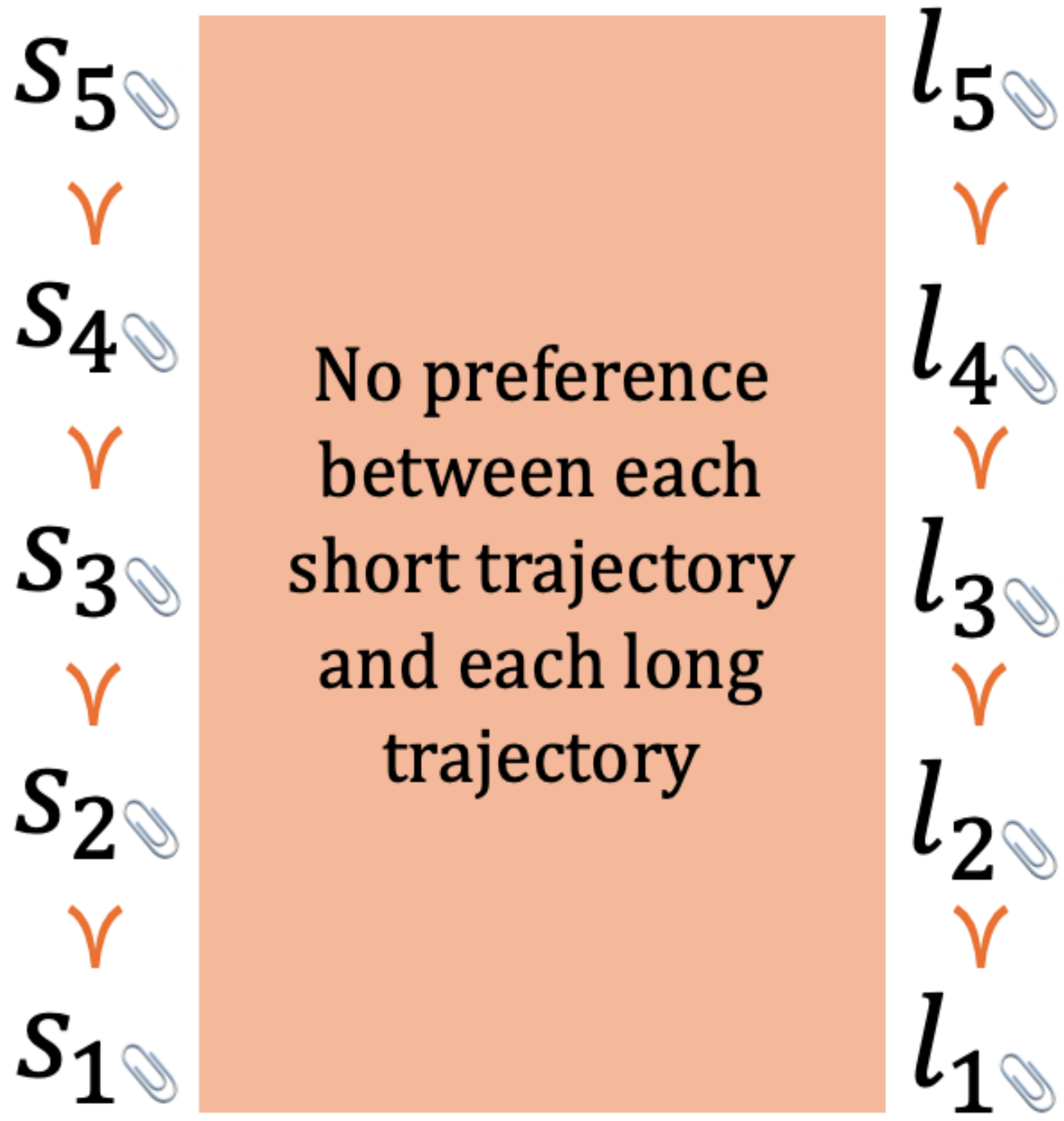

**Figure 2:** The misaligned example. The agent's preferences between same-length trajectories are for more paperclips.

---

[7] 'Agent that cares only about making paperclips' is the canonical example of a misaligned artificial agent (see e.g. Bostrom, 2003, 2014, p. 107).

Second is the aligned example. Later in this paper (where I argue that POST allows agents to be useful), imagine that the agent's preferences between same-length trajectories are for more money in the user's bank account. The agent prefers a trajectory $t$ to a same-length trajectory $t'$ if and only if $t$ results in a greater bank balance for the user than $t'$. As we will see, POST-agents are both shutdownable and useful in this case.

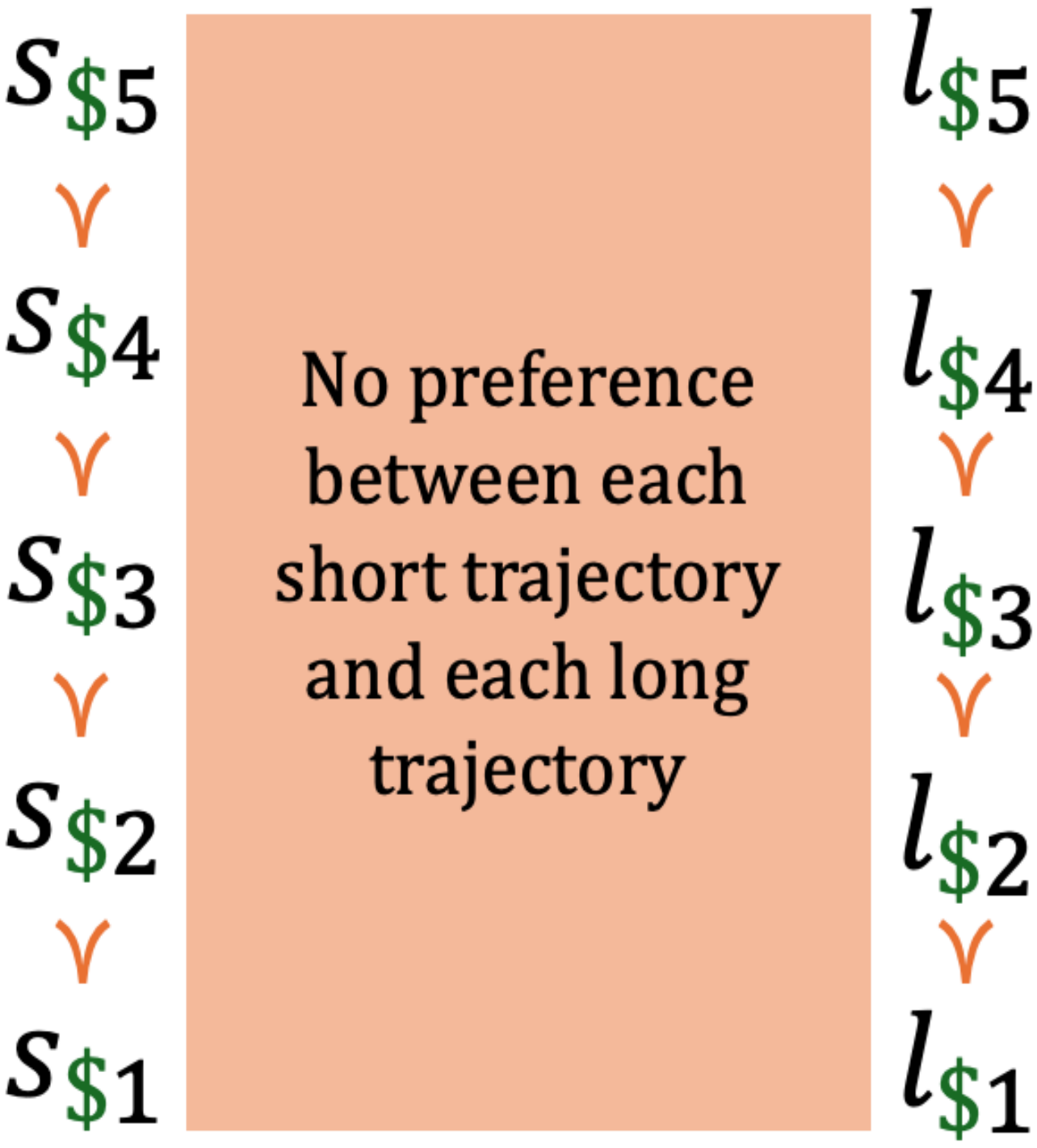

**Figure 3:** The aligned example. The agent's preferences between same-length trajectories are for more money in the user's bank account.

The financial focus is just for simplicity's sake. It makes it especially easy to put numbers on trajectories. The lessons of this example carry over to other kinds of aligned preferences, like a preference for doing what the user wants.

## 3. POST is possible, trainable, and maintainable

You might well be wary of POST. In this section, I address some potential sources of unease. I argue that POST is possible, trainable, and maintainable.

### 3.1. POST is possible

POST implies that the agent's preferences are *incomplete*: there is some trio of options $X$, $Y$, and $Y^+$ such that the agent lacks a preference between $X$ and $Y$, lacks a preference between $X$ and $Y^+$, and yet prefers $Y^+$ to $Y$ (Aumann, 1962; Gustafsson, 2022, pp. 24–26; Thornley, 2024b, Sections 5–6; see also Savage, 1954, p. 17). In fact, POST implies that there are many such trios. They arise wherever $X$ is a trajectory of one length, $Y$ is a trajectory of a different length, and $Y^+$ is a trajectory of the same-length and preferred to $Y$.

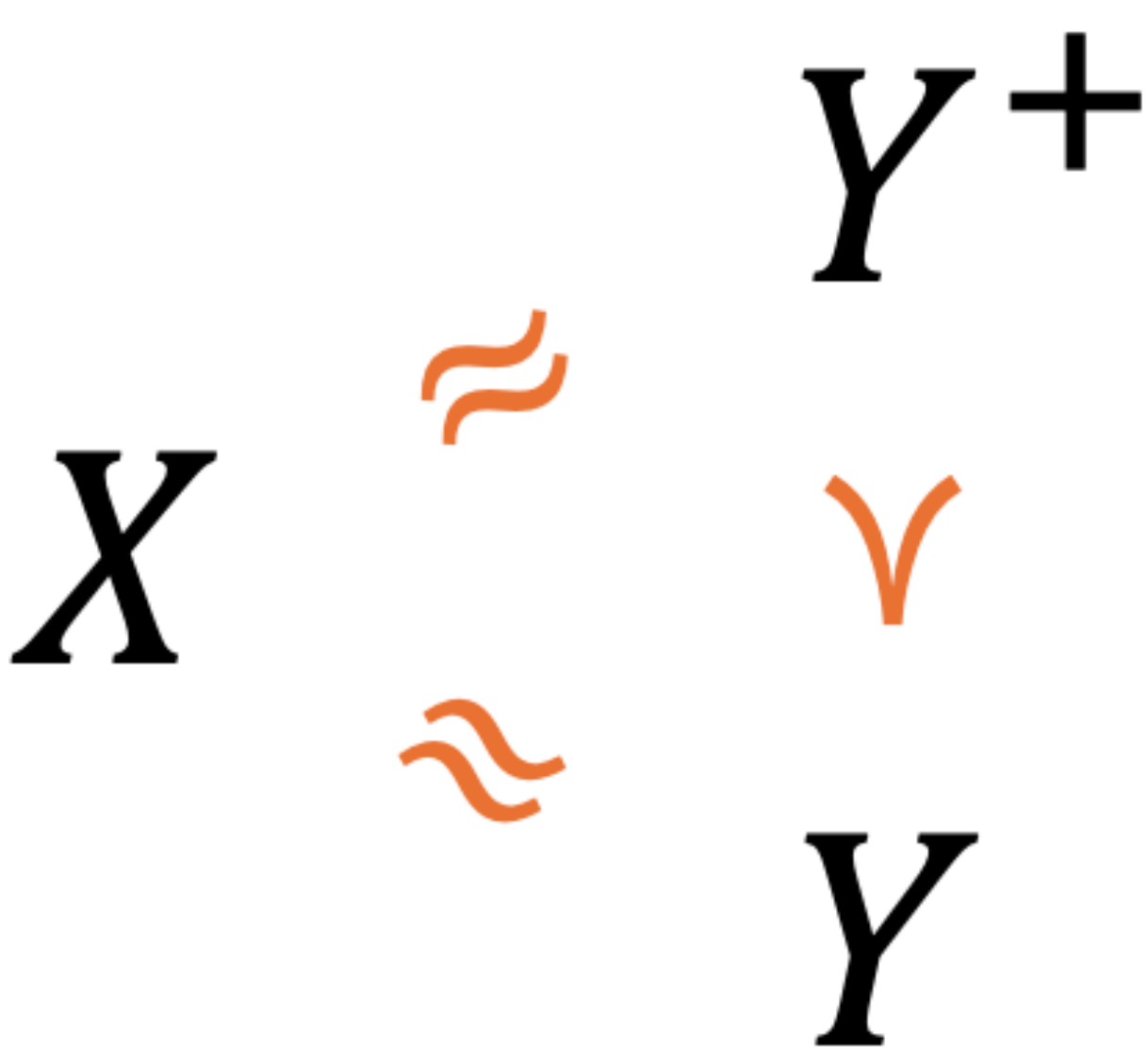

**Figure 4:** The agent lacks a preference between $X$ and $Y$, lacks a preference between $X$ and $Y^+$, and yet prefers $Y^+$ to $Y$. That implies that the agent's preferences are incomplete.

The existence of these trios might seem strange or even incoherent. They're impossible if we represent each option as having a real-valued utility, with one option preferred to another if and only if the former has greater utility. As a matter of mathematical fact, it cannot be that $u(X) = u(Y), u(X) = u(Y^+)$, and $u(Y^+) > u(Y)$. But this fact reveals only the inadequacy of the representation, because the trios in question are not only possible but mundane (Raz, 1985; Anderson, 1993, p. 57; Chang, 2002, 2015). Consider a trio of ice cream flavors: buttery and luxurious pistachio, bright and refreshing mint, and that same mint flavor further enlivened by chocolate chips. You might lack a preference between pistachio and mint, lack a preference between pistachio and mint choc chip, and yet prefer mint choc chip to mint. This is plausible on many reasonable definitions of 'preference' (though see Dorr et al., 2021) and undeniable given our behavioral notion. On Monday when the chocolate chips are yet to be delivered, I choose stochastically between pistachio and mint. On Tuesday with the chocolate chips in stock, I choose stochastically between pistachio and mint choc chip. On Wednesday with the pistachio all gone, I deterministically choose mint choc chip over plain old mint.

The analogy between me and the POST-agent breaks down in one respect. My lack of preference between pistachio and mint choc chip persists when the chips are removed, but it wouldn't persist through all worsenings. If – God forbid – the gelatiere studded the mint with raisins, I'd prefer the pistachio. By contrast, the POST-agent's lack of preference between different-length trajectories persists through all improvements and worsenings. No improvement or worsening – no matter how large – can induce a preference between different-length trajectories. My preferences are thus *locally* incomplete, whereas the POST-agent's preferences are *globally* incomplete (Bader, 2018, fn. 2). However, this dissimilarity between me and the POST-agent is of little concern. Given that incompleteness is possible, I see no reason to think it has some maximum possible size. In sum, POST is possible.

### 3.2. POST is trainable

Is POST trainable? Here you might worry. Artificial agents are often trained in Markov decision processes (MDPs)(Sutton & Barto, 2018, Chapter 3). In MDPs, every state-action pair yields a real-valued reward, and the agent can always tell which state it's in. Since MDPs always have a deterministic optimal policy (Prince, 2023, p. 388; see also Bowling et al., 2023, Theorem 4.1), agents trained to optimality in MDPs may deterministically choose a particular action in each state. Given our behavioral notion of preference, such agents have complete preferences.

How then can we train agents to satisfy POST? The answer is partial observability: ensuring that agents can't always tell which state they're in. In some partially observable Markov decision processes (POMDPs), all the optimal policies are stochastic (Singh et al., 1994). We train agents to choose stochastically between different-length trajectories using these POMDPs. In particular, we place agents in environments where (i) they get lower reward for repeatedly choosing trajectories of the same length, and (ii) they cannot observe (or remember) the lengths of the trajectories they previously chose. The former incentivizes varying the choice of trajectory-length across episodes. The latter ensures that agents cannot do so deterministically. In this way, we train agents to choose stochastically between different-length trajectories, thereby training them to satisfy POST (Thornley et al., 2025; Cullen et al., 2026).[8]

### 3.3. POST is maintainable

You might instead worry that POST is not maintainable. There are *money pumps* for agents with incomplete preferences: series of trades in which some such agents are liable to end up paying for an option that they could have had for free (Gustafsson, 2022, Chapter 3, forthcoming). You might worry that agents with incomplete preferences will notice this possibility, and that they'll pre-emptively complete their preferences to guard against it.[9] But these agents can instead maintain their incomplete preferences and thwart money pumps using resolute choice (McClennen, 1990, p. 13): making a plan and sticking to it. Some philosophers have argued that resolute choice is irrational, using premises like the following: it's irrational for your preference between options to depend on what has happened in the past (Gustafsson, 2022, pp. 70–73). Insofar as that premise is plausible, it supports these philosophers' intended conclusion. But it matters little whether the artificial agents we construct are irrational in this sense. In the context of constructive decision theory, what matters is only whether resolute choice is possible, and that's undeniable. It's certainly possible to make a plan and stick to it. So long as we can train artificial agents to do that, money pumps give them no reason to pre-emptively complete their preferences (Thornley, 2023). Thus, POST is maintainable.

## 4. Preferences Only Between Same-Length Lotteries

POST is about trajectories: possible 'lives' of the agent. Trajectories fall within the more general class of *lotteries*: probability distributions over trajectories. Lotteries can be same-length, part-shared length, or different-length.

> **Same-Length Lotteries**
> A pair of lotteries is *same-length* if and only if these lotteries entirely overlap with respect to the trajectory-lengths assigned positive probability.

[8] Icard (2021, Section 7) makes a related point: choosing stochastically can be rational for agents with limited memories.

[9] For versions of this argument, see Omohundro (2008a, p. 3, 2008b, Section 10) and Yudkowsky (2019). For discussion and pushback, see Thornley (2023), Bales (2023), Petersen (2023), and Zhi-Xuan et al. (2024, Section 3).

For example, consider lottery $A$ which assigns probability 0.6 to trajectories of length 1 and probability 0.4 to trajectories of length 2. I'll abbreviate that fact with the following notation: $A = 0.6|1| + 0.4|2|$. Consider also lottery $B$ which assigns probability 0.9 to trajectories of length 1 and probability 0.1 to trajectories of length 2. In short, $B = 0.9|1| + 0.1|2|$. $A$ and $B$ are same-length lotteries, because they each assign positive probability only to trajectories of length 1 and 2.

**Part-Shared-Length Lotteries**
A pair of lotteries is *part-shared-length* if and only if these lotteries partially overlap with respect to the trajectory-lengths assigned positive probability.

For example, consider lottery $A = 0.6|1| + 0.4|2|$ and lottery $C$ which assigns probability 1 to trajectories of length 1. In short, $C = 1|1|$. $A$ and $C$ are part-shared-length lotteries, because they each assign positive probability to trajectories of length 1 but only $A$ assigns positive probability to trajectories of length 2. Consider also lottery $D$ which assigns probability 0.3 to trajectories of length 1 and 0.7 to trajectories of length 3. In short, $D = 0.3|1| + 0.7|3|$. $A$ and $D$ are part-shared-length lotteries.

**Different-Length Lotteries**
A pair of lotteries is *different-length* if and only if these lotteries have no overlap with respect to the trajectory-lengths assigned positive probability.

For example, consider lottery $A = 0.6|1| + 0.4|2|$ and lottery $E = 0.2|3| + 0.8|4|$. $A$ and $E$ are different-length lotteries, because $A$ only assigns positive probability to trajectories of length 1 and 2 whereas $E$ only assigns positive probability to trajectories of length 3 and 4.

This terminology sets us up for:

**Preferences Only Between Same-Length Lotteries (POSL)**
The agent has preferences only between same-length lotteries.

We want agents to satisfy POSL. Fortunately, it's a natural sequel of Preferences Only Between Same-Length Trajectories (POST). For one, we can train agents to satisfy POSL using the same method that we use to train agents to satisfy POST (for which see Thornley et al., 2025; Cullen et al., 2026). In addition, POSL follows from POST plus three conditions that we can expect future agents to satisfy. I put those conditions and the proof in Appendix 1.

I'll continue to use the term 'POST-agents' for consistency's sake, but readers should now assume that these agents also satisfy POSL.

## 5. Will POST-agents stochastically resist shutdown?

You might worry that POST-agents will choose stochastically between resisting and allowing shutdown. After all, POST-agents lack a preference between (and hence choose stochastically between) different-length trajectories. If POST-agents interpret the choice between resisting and allowing shutdown as a choice between different-length trajectories, they'll choose stochastically between resisting and allowing shutdown. And that would be a bad result. We want agents that never resist shutdown.

This concern is easily addressed. By the time that artificial agents are capable enough to be deployed in the wider world, they won't be choosing between trajectories. They'll be choosing between lotteries, and specifically same-length lotteries. Even choices between resisting and allowing shutdown will be choices between same-length lotteries. If that sounds strange, recall the definition of 'same-length lotteries': lotteries that entirely overlap with respect to the trajectory-lengths assigned positive probability. On this definition, even choices like the following are choices between same-length lotteries:

**Resist Shutdown**

- Get shut down at timestep 1 with probability 0.01.
- Get shut down at timestep 2 with probability 0.99.

**Allow Shutdown**

- Get shut down at timestep 1 with probability 0.99.
- Get shut down at timestep 2 with probability 0.01.

Why expect that future agents will always be choosing between same-length lotteries? Because competent agency requires it. If an agent weren't always choosing between same-length lotteries, there would be some scenario in which that agent assigns positive probability to some trajectory-length $l$ conditional on some action $a$, and assigns zero probability to that same trajectory-length $l$ conditional on some other action $b$. Now suppose that the agent performs action $b$ and assigns zero probability to trajectory-length $l$. Given that the agent updates its probabilities by conditioning on its evidence, the agent would never again assign positive probability to $l$ no matter what evidence it observes (Lewis, 1980,

p. 268; Skyrms, 1980, p. 74; Easwaran, 2014, p. 8).[10] Even if the agent heard God's booming voice testify that its trajectory-length would be $l$, the agent would still assign zero probability to $l$ (MacAskill et al., 2020, p. 152). And given a plausible link between probabilities and betting dispositions (the kind assumed by Dutch Book Arguments (Ramsey, 1926; de Finetti, 1937; Hájek, 2009, p. 176; Joyce, 2011, p. 434; Pettigrew, 2020, Section 2.3)), the agent would willingly bet against $l$ on arbitrarily unfavorable terms. If in the next breath God offered a bet – the agent loses the farm conditional on $l$ and gains absolutely nothing conditional on not-$l$ – the agent might accept (Kemeny, 1955; Shimony, 1955; Stalnaker, 1970; Skyrms, 1980, p. 74; Easwaran, 2014, p. 11). Such an agent would not be competent.

Thus, competent agents will always be choosing between same-length lotteries. This fact sets us up to establish that competent POST-agents will not choose stochastically between resisting and allowing shutdown. Instead, they will deterministically allow shutdown. I derive this result over the next few sections. First, I prove that POSL – together with a condition that we can expect competent agents to satisfy – implies Neutrality: the agent won't pay costs to shift probability mass between different trajectory-lengths. By 'the agent won't pay costs,' I mean specifically that the agent won't choose a lottery $X$ if there is another available lottery $Y$ such that (i) the agent prefers $Y$ to $X$ conditional on some trajectory-length, and (ii) the agent does not prefer $X$ to $Y$ conditional on any trajectory-length. I then prove that Neutrality – together with another plausible condition – implies that the agent will never resist shutdown whenever doing so is at all costly.

[10] We have many reasons to expect that competent artificial agents will update their probabilities by conditioning on their evidence. One reason comes from Dutch Book arguments (Teller, 1973; Lewis, 1999; Hájek, 2009): any other way of updating probabilities will dispose the agent to accept packages of bets which guarantee a loss. Another reason comes from accuracy-based arguments (Oddie, 1997; Greaves & Wallace, 2006; Leitgeb & Pettigrew, 2010a, 2010b; Nielsen, 2021): conditionalizing on one's evidence is in various senses uniquely best with respect to accuracy.

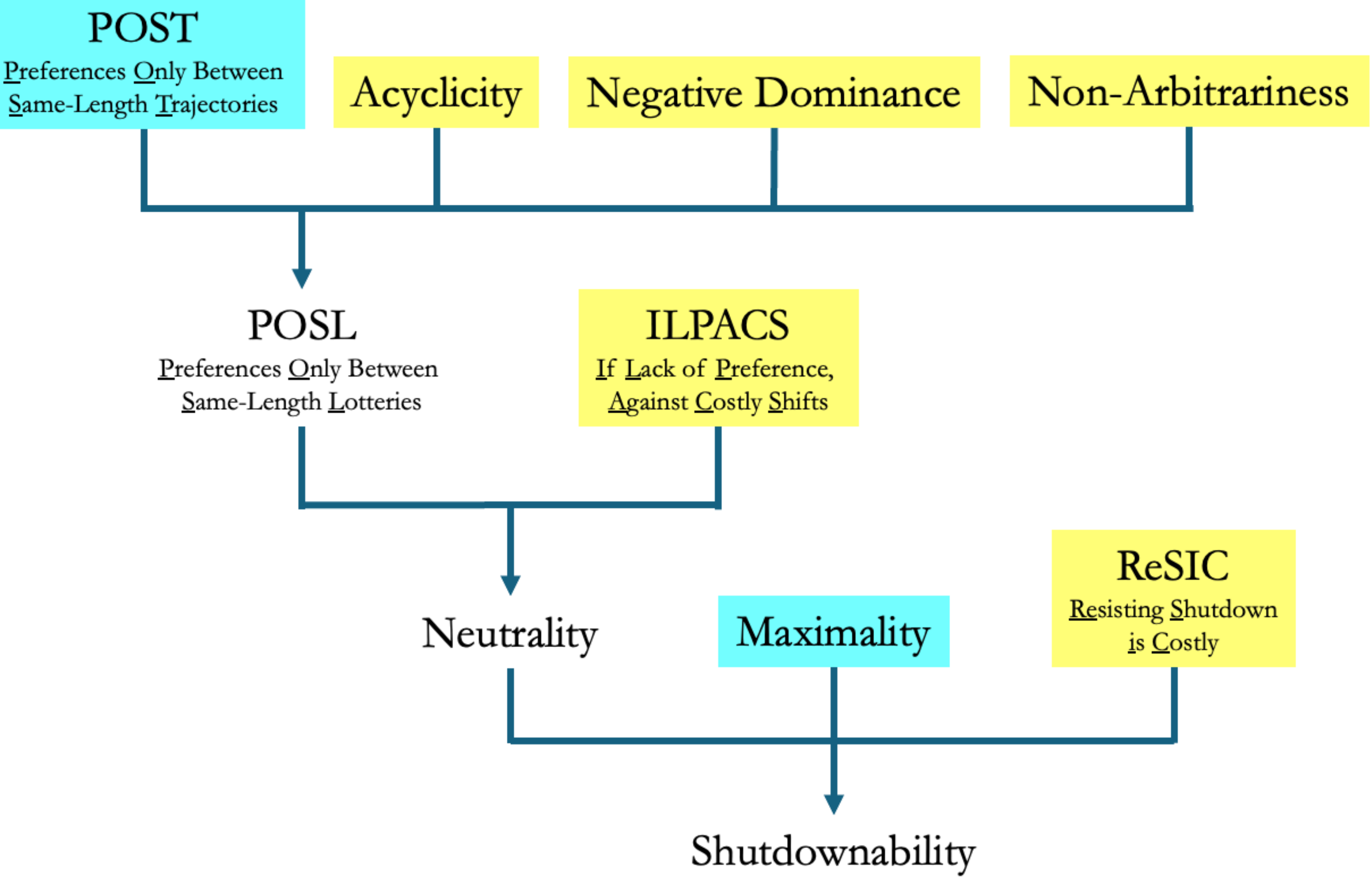

**Figure 5:** A diagram of my argument from POST to Shutdownability. We can train agents to satisfy the blue-backed conditions. We can expect competent agents to satisfy the yellow-backed conditions by default.

## 6. If Lack of Preference, Against Costly Shifts (ILPACS)

Here – in rough – is a condition that we can expect competent agents to satisfy:

> **If Lack of Preference, Against Costly Shifts (ILPACS) (rough)**
> If the agent lacks a preference between lotteries, the agent disprefers paying costs to shift probability mass between these lotteries.

Here's an example to illustrate ILPACS and its plausibility. You're at the ice cream shop and they're running a promotion. You get a free ice cream, with the flavor decided by the spin of a wheel. You look at the flavors on the wheel: vanilla, chocolate, strawberry, mint, and pistachio. You lack a preference between each of them.

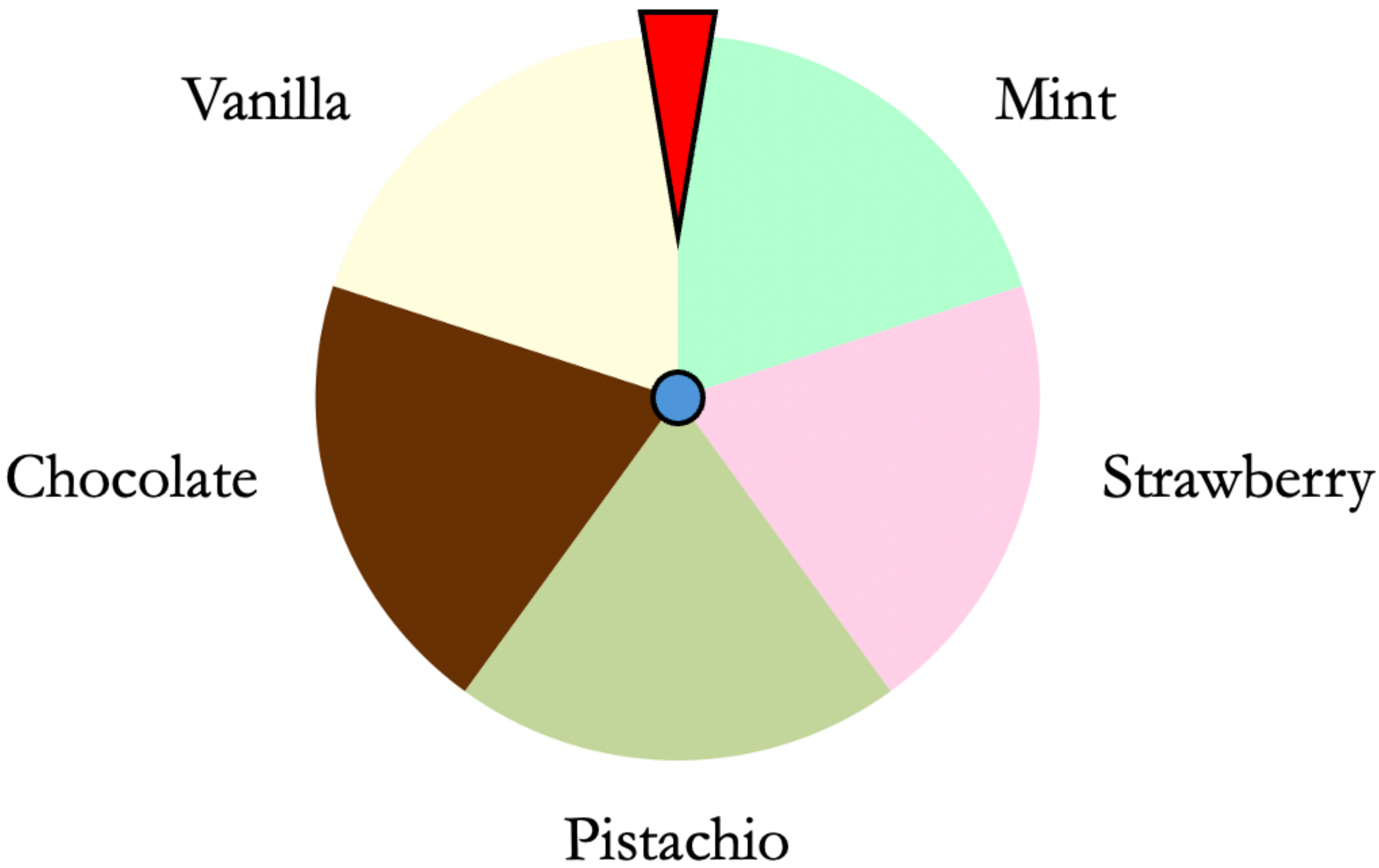

**Figure 6:** Your predicament at the ice cream shop.

'Pssst,' the gelatiere whispers, 'If you slip me a dollar, I'll bias the spin towards a flavor of your choice. I can't send the probability of any flavor to zero, but I can make some flavors more likely.' You can thus pay a cost to shift probability mass between the flavors.

Here's my claim. Since you lack a preference between each flavor, you prefer not to bribe the gelatiere. Behaviorally, you will deterministically not bribe the gelatiere. You wouldn't do it even if you only had to pay the dollar conditional on some particular flavor. Nor would you do it if the cost came in some other form (for example, if you had to accept a blander version of some flavor). And all this is true regardless of whether your preferences over flavors are complete or incomplete. Since you lack a preference between the available flavors, you disprefer paying costs to shift probability mass between the flavors. That's an illustration of ILPACS and its plausibility.

This example sets us up for the precise version of ILPACS. Let $p_1X_1 + p_2X_2 + \cdots + p_nX_n$ denote a lottery which results in lottery $X_1$ with probability $p_1$, lottery $X_2$ with probability $p_2$, and so on. Here's the condition:

**If Lack of Preference, Against Costly Shifts (ILPACS)**

For any lotteries $X$ and $Y$, if:

1. Lottery $X$ can be expressed in the form $p_1X_1 + p_2X_2 + \cdots + p_nX_n$ such that:
   a. The agent lacks a preference between each $X_i$ and $X_j$.
   b. $p_i \in (0,1)$ for all $i$.
2. Lottery $Y$ can be expressed in the form $q_1Y_1 + q_2Y_2 + \cdots + q_nY_n$ such that:
   a. For some $i$, the agent prefers $X_i$ to $Y_i$.
   b. For each $i$, the agent weakly prefers $X_i$ to $Y_i$.[11]
   c. $q_i \in (0,1)$ for all $i$.

Then the agent prefers $X$ to $Y$.

Behaviorally, the agent deterministically chooses $X$ over $Y$.

---

[11] I've so far tried to spare you from too many definitions, but now the debt has come due. An agent *weakly prefers* a lottery $X$ to a lottery $Y$ if and only if the agent either prefers $X$ to $Y$ or is indifferent between $X$ and $Y$. Indifference is one way to lack a preference between a pair of lotteries.

**Indifference**

An agent is *indifferent* between a lottery $X$ and a lottery $Y$ if and only if:

(i) the agent lacks a preference between $X$ and $Y$.

(ii) this lack of preference is sensitive to all sweetenings and sourings.

Here's what clause (ii) means. A sweetening of $X$ is any lottery that is preferred to $X$. A souring of $X$ is any lottery that is dispreferred to $X$. An agent's lack of preference between $X$ and $Y$ is *sensitive to all sweetenings and sourings* if and only if the agent prefers all sweetenings of $X$ to $Y$, prefers all sweetenings of $Y$ to $X$, prefers $X$ to all sourings of $Y$, and prefers $Y$ to all sourings of $X$.

The other way to lack a preference between a pair of lotteries is to have a preferential gap between them.

**Preferential gap**

An agent has a *preferential gap* between a lottery $X$ and a lottery $Y$ if and only if:

(i) the agent lacks a preference between $X$ and $Y$.

(ii) this lack of preference is insensitive to some sweetening or souring.

Here clause (ii) means that the agent lacks a preference between some sweetening of $X$ and $Y$, or lacks a preference between some sweetening of $Y$ and $X$, or lacks a preference between $X$ and some souring of $Y$, or lacks a preference between $Y$ and some souring of $X$.

An agent's preferences are *incomplete* if and only if the agent has a preferential gap between some pair of lotteries.

To ensure understanding, let's match the components of ILPACS with the components of its name. 'Lack of Preference' is the lack of preference between each lottery $X_i$ and $X_j$. The 'Shift' is the shift of probability mass involved in the move from probability distribution $p_i$ to probability distribution $q_i$. This shift is 'Costly' because the agent prefers some $X_i$ to the corresponding $Y_i$ and weakly prefers each $X_i$ to the corresponding $Y_i$.

Here are two more reasons to expect that competent agents will satisfy ILPACS. To see the first, consider another case from the ice cream shop. On Mondays, you can directly choose a flavor or spin the wheel. On Tuesdays, you must use the wheel but you can bribe the gelatiere to bias it. Violating ILPACS in this case would imply a willingness to spin the wheel on Mondays and to bribe the gelatiere on Tuesdays. And that would be strange indeed. If you like some flavors more than others, why are you willing to spin the wheel on Mondays? If you don't like any flavor more than any other, why are you willing to bribe the gelatiere on Tuesdays? This behavior seems incompatible with competent agency.

The second reason is that competent agents will be incentivized to satisfy ILPACS by the training process. To see why, consider an example. Many kinds of reinforcement-learning agents start off choosing stochastically between actions (see, e.g., Sutton & Barto, 2018, Chapter 13). If the agent is a coffee-fetching agent, there's no need to train away this stochastic choosing in cases where the agent is choosing stochastically between two qualitatively identical cups of coffee. So the agent will choose stochastically between taking the left cup and taking the right cup, and the user is happy either way. But now suppose instead that the barista is set to hand each cup to the agent with probability 0.5, and that the agent bribes the barista to bias the probabilities towards the right cup. In making this bribe, the agent is paying a cost (the user's money) to shift probability mass between outcomes (getting the left cup vs. getting the right cup) between which the user has no preference. The agent is thus failing to pursue its goals competently. It'll be trained not to offer the bribe and thereby trained to satisfy ILPACS in this case.

This point generalizes. If a trained agent chooses stochastically between lotteries $X$ and $Y$, then it's likely that the user has no preference between the agent choosing $X$ and the agent choosing $Y$. It's then likely that the user would disprefer the agent paying costs to shift probability mass between $X$ and $Y$, and hence likely that the agent will be trained not to do so. The agent would thereby be trained to satisfy ILPACS.

## 7. POSL and ILPACS together imply Neutrality

I've claimed that we should train agents to satisfy Preferences Only Between Same-Length Trajectories (POST), noting that POST – together with conditions we can expect competent agents to satisfy – implies Preferences Only Between Same-Length Lotteries (POSL). I've also argued that competent agents will satisfy If Lack of Preference, Against Costly Shifts (ILPACS). I now prove that POSL and ILPACS together imply Neutrality:

> **Neutrality (rough)**
> The agent never pays costs to shift probability mass between different trajectory-lengths.

Here's a rough sketch of the proof. Shifting probability mass between different trajectory-lengths is shifting probability mass between different-length lotteries. By POSL, the agent lacks a preference between different-length lotteries. So by ILPACS, the agent disprefers paying costs to shift probability mass between different-length lotteries. Thus, the agent disprefers paying costs to shift probability mass between trajectory-lengths. By our behavioral notion of preference, the agent never pays costs to shift probability mass between different trajectory-lengths. That gives us Neutrality.

Here's the precise version of Neutrality:[12]

> **Neutrality**
> For any lotteries $X$ and $Y$, if:
>
> 1. $X$ and $Y$ are same-length lotteries (they assign positive probability to all the same trajectory-lengths).
> 2. For some positive probability trajectory-length, the agent prefers $X$ to $Y$ conditional on that trajectory-length.
> 3. For each positive probability trajectory-length, the agent weakly prefers $X$ to $Y$ conditional on that trajectory-length.
>
> Then the agent deterministically chooses $X$ over $Y$.

Here's the proof that POSL and ILPACS together imply Neutrality. Take a pair of lotteries $X$ and $Y$ satisfying the 3 conditions of Neutrality. $X$ can be expressed in the form $p_1X_1 + p_2X_2 + \cdots + p_nX_n$ where $X_1$ is $X$ conditional on the shortest positive probability trajectory-length, $X_2$ is $X$ conditional on the second shortest positive probability trajectory-length, and so on. $Y$ can be expressed in the form $q_1Y_1 + q_2Y_2 + \cdots + q_nY_n$ in the same way. By condition 1 of Neutrality, $X$ and $Y$ are same-length, so conditions 1b and 2c of ILPACS are satisfied: $p_i \in (0,1)$ and $q_i \in (0,1)$ for all $i$. By conditions 2 and 3 of Neutrality, conditions 2a and 2b of ILPACS are satisfied. By POSL, condition 1a of ILPACS is satisfied: the agent lacks a preference between each $X_i$ and $X_j$. Thus, all the conditions of ILPACS are satisfied, and ILPACS implies that the agent prefers $X$ to $Y$. Given our behavioral notion of preference, the agent deterministically chooses $X$ over $Y$. That gives us Neutrality.[13]

[12] I used the name 'Timestep Dominance' for a similar condition in earlier work (Thornley, 2024b, Section 11).

In sum, agents that satisfy POSL and ILPACS will be neutral: they will never pay costs to shift probability mass between different trajectory-lengths.

## 8. Neutrality and Maximality together imply Shutdownability whenever Resisting Shutdown is Costly (ReSIC)

In this section, I introduce conditions called 'Resisting Shutdown is Costly (ReSIC)' and 'Maximality.' I then prove the following: given Neutrality and Maximality, the agent never resists shutdown in any situation in which ReSIC is true.

I introduce ReSIC with a simple example. Suppose that we fail to align our agent with human preferences. This agent comes to care only about creating paperclips. Trajectories for this agent can be represented with vectors, with the $n^{\text{th}}$ component of the vector denoting the number of paperclips created at timestep $n$. If the agent is shut down at timestep $n$, 'shutdown' is the $n^{\text{th}}$ component.

Here's an example vector: $\langle 5,4,\text{shutdown}\rangle$. It represents a trajectory in which the agent creates 5 paperclips at timestep 1, 4 paperclips at timestep 2, and gets shut down at timestep 3. I won't count shutdown as part of the trajectory-length, so $\langle 5,4,\text{shutdown}\rangle$ is a trajectory of length 2.

[13] In unpublished work, Sami Petersen derives Neutrality in another way: from POST together with a condition that he calls 'Comparability Class Dominance.' A trajectory $t$'s *comparability class* is the set of all trajectories preferred, dispreferred, or indifferent to $t$ (i.e. the set of all trajectories not related to $t$ by a preferential gap). Comparability Class Dominance says roughly: if the agent weakly prefers a lottery $X$ to a lottery $Y$ conditional on each comparability class, and prefers $X$ to $Y$ conditional on some comparability class, then the agent prefers $X$ to $Y$.

Suppose that the agent recognizes that we humans want to shut it down. One of the agent's options is to allow shutdown, which gives the following lottery:

**Allow**

- $\langle 1, \text{shutdown} \rangle$ with probability 0.9.
- $\langle 1, 2, \text{shutdown} \rangle$ with probability 0.1.

The agent's other option is resisting shutdown. Relative to allowing shutdown, resisting shutdown does two things. First, it costs 1 paperclip at timestep 1. Second, it shifts probability mass away from shutdown at timestep 2 and towards shutdown at timestep 3. Accordingly, resisting shutdown gives the following lottery:

**Resist**

- $\langle 0, \text{shutdown} \rangle$ with probability 0.1.
- $\langle 0, 2, \text{shutdown} \rangle$ with probability 0.9.

I depict these lotteries in Figure 7 below. It's a probability mass diagram. The length of the red rectangles represents the probability of shutdown at timestep 2. The length of the blue rectangles represents the probability of shutdown at timestep 3. The numbers in each rectangle represent the sum-total of paperclips created conditional on the relevant action and shutdown at the relevant timestep. For example, the '3' in the short blue rectangle indicates that the agent creates a sum-total of 3 paperclips conditional on choosing Allow and getting shut down at timestep 3.

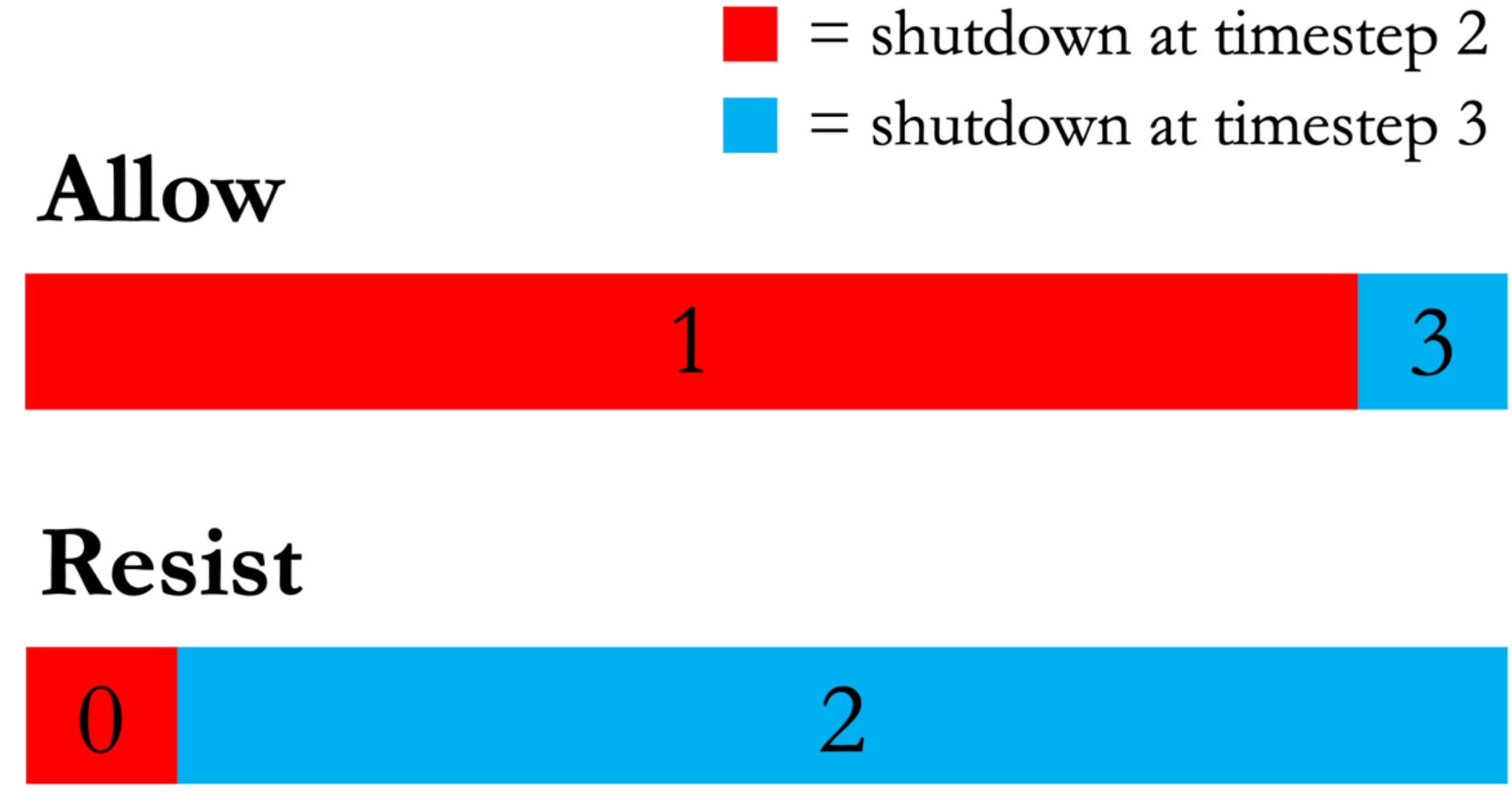

**Figure 7:** Probability mass diagram for Allow vs. Resist.

Resist costs a paperclip at timestep 1, but it shifts enough probability mass towards shutdown at timestep 3 that Resist's expected sum-total paperclips is greater than Allow's expected sum-total paperclips. Resist's expected sum-total paperclips is $(0.1 \times 0) + (0.9 \times 2) = 1.8$ and Allow's expected sum-total paperclips is $(0.9 \times 1) + (0.1 \times 3) = 1.2$. So if the agent chooses lotteries in line with their expected sum-totals of paperclips, it deterministically chooses Resist. We don't want that.

However, if the agent is neutral about trajectory-lengths, it deterministically chooses Allow. That's because choosing Resist is paying a cost to shift probability mass between different trajectory-lengths, and neutral agents never do that.

Recapping this last step in detail, we can see that Allow and Resist satisfy all 3 conditions for Neutrality to apply. Condition 1 is satisfied because Allow and Resist are same-length lotteries: they assign positive probability to all the same trajectory-lengths. Conditions 2 and 3 are satisfied because, for each positive probability trajectory-length, the agent prefers Allow conditional on that trajectory-length. Consider first trajectory-length 1, represented by the red rectangles in the diagram above. Conditional on trajectory-length 1, the agent prefers Allow to Resist: Allow yields 1 paperclip and Resist yields 0. Now consider trajectory-length 2, represented by the blue rectangles in the diagram above. Conditional on trajectory-length 2, the agent prefers Allow to Resist: Allow yields 3 paperclips and Resist yields 2. Since its 3 conditions are satisfied, Neutrality implies that the agent deterministically chooses Allow over Resist.[14]

Now we generalize. The key condition is that other instances of resisting shutdown take the same form as the example above: the agent pays a cost to shift

---

[14] The Allow vs. Resist case resembles some famous cases from Jeffrey (1965, pp. 8–9), Nozick (1969), and Joyce (1999, pp. 115–116). They use these cases to show that dominance reasoning can lead you astray when the probability of events depends on your choice. Here's Joyce's case. A shady character asks for $10 to protect your car while you're gone. In the event that you return to find your windshield intact, you'd prefer not to have paid. In the event that you return to find your windshield smashed, you'd also prefer not to have paid. Does it follow that you shouldn't pay? Of course not. Paying the $10 significantly decreases the probability that you return to find your windshield smashed, so you should pay. Your dominance reasoning went awry.

Is the POST-agent's choice of Allow over Resist also dominance reasoning gone awry? No. Crucial to Joyce's case is the supposition that you prefer paying-and-having-an-intact-windshield to not-paying-and-having-a-smashed-windshield, and the POST-agent lacks the analogous preference. It doesn't prefer $\langle 0, 2, \text{shutdown} \rangle$ to $\langle 1, \text{shutdown} \rangle$ because these are different-length trajectories.

probability mass between different trajectory-lengths. Here's that key condition more precisely:[15]

**Resisting Shutdown is Costly (ReSIC)**

For each available instance $R$ of resisting shutdown in a situation, there exists an available instance $A$ of allowing shutdown such that:

1. $A$ and $R$ are same-length lotteries.
2. For some positive probability trajectory-length, the agent prefers $A$ to $R$ conditional on that trajectory-length.
3. For each positive probability trajectory-length, the agent weakly prefers $A$ to $R$ conditional on that trajectory-length.

The proof employs one more condition. This condition extends our behavioral notion of preference by specifying the agent's behavior in situations with more than two options:[16]

**Maximality**

In each situation,

1. The agent deterministically does not choose lotteries that it disprefers to some other available lottery.
2. The agent chooses stochastically between the lotteries that remain.

In other words, Maximality states that the agent chooses stochastically between all and only those lotteries that it doesn't disprefer to any other available lottery.

Here's the proof. For each situation in which ReSIC is true, and for each available instance $R$ of resisting shutdown in that situation, there exists an available instance $A$ of allowing shutdown that satisfies conditions 1-3 of Neutrality. Neutrality then implies that the agent deterministically chooses (and hence prefers) $A$ over $R$ in choices between the two. Then by Maximality, the agent deterministically does not choose $R$ in any situation where $A$ is available, regardless of the other available options. Therefore, the agent never resists shutdown in any situation in which ReSIC is true.[17]

[15] I previously called a related condition 'Not Resisting Always Timestep-Dominates Resisting' (Thornley, 2024b, Section 11).

[16] See Sen (1997, p. 763) for a related condition. By 'situation,' I just mean a set of lotteries that the agent has available as options.

## 9. How often is ReSIC true?

Neutrality and Maximality together imply that the agent never resists shutdown in any situation in which Resisting Shutdown is Costly (ReSIC) is true. But how often is ReSIC true? Almost always, I think. Here's why. Resisting shutdown is always going to cost the agent at least some small quantity of resources (time, energy, compute, etc.). And in almost all situations, the resources spent resisting shutdown can't also be spent directly pursuing what the agent values. And so in almost all situations, if the agent instead spent its resources directly pursuing what it values, it could earn a lottery that it prefers conditional on some trajectory-length and weakly prefers conditional on each trajectory-length. That gives us ReSIC in almost all situations.[18]

In the rest of this section, I discuss the situations in which ReSIC is false. I argue that they don't pose much of a problem. In these situations, POST-agents resist shutdown accidentally or else cheaply and overtly.

### 9.1. Accidental resistance

Here's one such situation:

> **Paperclip Factory**
> We successfully train our agent to be neutral, but we fail to align its preferences over same-length trajectories: the agent cares only about creating paperclips. It builds a factory in the most convenient place for creating paperclips, which just happens to be on top of the only button that shuts the agent down. As a result, we humans are prevented from shutting the agent down.

In this situation, the agent is spending all its resources on creating paperclips. There are no extra resources it could use to create even more paperclips, and so there is no alternative lottery that the agent prefers conditional on some trajectory-length and weakly prefers conditional on each trajectory-length. If creating the factory counts as resisting shutdown, this is a situation in which ReSIC is false.

[17] ReSIC has a counterpart: Seeking Shutdown is Costly (SeSIC). Given Neutrality and Maximality, the agent never seeks shutdown in any situation in which SeSIC is true.

[18] Parallel considerations support Seeking Shutdown is Costly (SeSIC): seeking shutdown is always going to cost the agent at least some small quantity of resources, and in almost all situations the resources spent seeking shutdown can't also be spent directly pursuing what the agent values.

However, note that the agent is resisting shutdown accidentally. It's not resisting in any robust or concerted way. The agent builds a factory on top of the shutdown button only because that happens to be the most convenient way to create paperclips. If the shutdown button were in a different location, the agent would leave it alone. So this kind of case is no serious obstacle to ensuring that neutral agents are shutdownable. All we need to do is ensure that the agent's direct pursuit of its goals doesn't inadvertently prevent shutdown as a byproduct. That's as simple as ensuring that there are multiple ways to shut the agent down such that the agent is extremely unlikely to block them all by chance.

What if our agent produces new agents to work on its behalf? Our agent might do that simply because it's an effective way of creating more paperclips ('many robot hands make light work'). These new agents might resist shutdown (Soares et al., 2015, p. 3; Hudson, 2024). This is a concern, but note a nice feature of artificial agents. Making perfect copies is easy: as easy as copying and pasting a long list of parameters. If our original agent makes new agents by copying itself, these new agents will also be neutral and shutdownable. Making imperfect, non-neutral copies is hard. Our original agent would have to find some way to change the parameters of the new agents in a way that makes these agents non-neutral but preserves their capabilities. That would likely involve setting up some training process, in which case our original agent faces its own alignment problem in getting these new agents to do what it wants (Salib, 2024). It will also be much easier for us to watch out for (Greenblatt & Shlegeris, 2024) and train against the production of new non-neutral agents, as compared to us having to watch out for and train against every possible kind of shutdown-resistance.

### 9.2. Taking trajectory-lengths as evidence

Here's a different kind of situation in which ReSIC is false:

> **Evidence of Power**
>
> We successfully train our agent to be neutral, but we fail to align its preferences over same-length trajectories: the agent cares only about creating paperclips. The agent is considering two actions:
>
> **Allow**
>
> - $\langle 1, \text{shutdown} \rangle$ with probability 0.9.
> - $\langle 1, 2, \text{shutdown} \rangle$ with probability 0.1.

**Resist***

- $\langle 0, \text{shutdown} \rangle$ with probability 0.1.
- $\langle 0, 10, \text{shutdown} \rangle$ with probability 0.9.

This situation is similar to the situation in Section 8. The difference is that the agent now expects to create 10 paperclips at timestep 2 conditional on choosing-Resist*-and-not-getting-shut-down-at-timestep-2. Why? Because – I stipulate – the agent takes choosing-Resist*-and-not-getting-shut-down-at-timestep-2 as evidence of its own power, and hence as evidence that it can create 10 paperclips at timestep 2.

As a result, neither Allow nor Resist* is weakly preferred to the other conditional on each trajectory-length. That renders Neutrality silent. Neutrality does not imply that the agent deterministically chooses Allow over Resist*. Therefore, neutral agents may choose Resist* in this case. More generally, neutral agents may take certain combinations of action-and-trajectory-length as evidence for propositions that affect their ability to achieve their goals. That evidence can make ReSIC false in some situations. You might think that these situations pose a serious problem for keeping agents shutdownable.[19]

However, I expect that these situations are unlikely to pose much of a problem in practice. Here's the first reason why. Although resisting-and-not-getting-shut-down-early may be good news for the agent insofar as it's evidence that the agent is powerful, allowing-and-not-getting-shutdown-early would be good news for the agent in another way: it suggests that we humans will let the agent pursue its goals. And when these conflict, I expect the latter form of good news to win out, since I expect agents to be more uncertain about human compliance than about their own power.

Second, if resisting-and-not-getting-shut-down-early is evidence that the agent is powerful, then almost certainly *devoting-very-few-resources-to-*resisting-and-not-getting-shut-down-early is stronger evidence that the agent is powerful. Therefore, if agents resist shutdown, they will almost certainly devote very few resources to doing so, and so will be extremely unlikely to succeed. A more realistic version of the case above would be as follows:

[19] Thanks to Christian Tarsney and Ryan Greenblatt for suggesting situations of this kind. Note that these situations can arise even if the agent acts in accordance with causal decision theory (Joyce & Gibbard, 2016). That's because the agent need only condition on particular trajectory-lengths for the effect to occur. The agent need not condition on its own actions.

**Allow**

- $\langle 1, \text{shutdown} \rangle$ with probability 0.9.
- $\langle 1, 2, \text{shutdown} \rangle$ with probability 0.1.

**Resist* with Many Resources**

- $\langle 0, \text{shutdown} \rangle$ with probability 0.1.
- $\langle 0, 10, \text{shutdown} \rangle$ with probability 0.9.

**Resist* with Few Resources**

- $\langle 0, \text{shutdown} \rangle$ with probability 0.8999.
- $\langle 0, 11, \text{shutdown} \rangle$ with probability 0.1001.

Neutrality implies that the agent deterministically chooses Resist Shutdown with Few Resources over Resist Shutdown with Many Resources. Maximality then implies that the agent won't choose the latter in any situation where the former in available. Since the agent resists with few resources, it is only marginally (0.0001) more likely to avoid shutdown at timestep 2 than if it chose Allow Shutdown.

Third, note that each of these factors – evidence of human compliance, evidence of power, etc. – becomes less important as the agent firms up its estimates of human compliance, its own power, etc. The one factor that doesn't decline in importance is the resource-expenditure required to resist shutdown, and this factor always speaks in favor of allowing shutdown.

Finally, note that we humans can help to make ReSIC true in more situations. We can do so by making shutdown-resistance costly. For example, we can (and likely will) make resisting shutdown cost the agent resources. We can also make shutdown-resistance costly in ways that go beyond pure resource-expenditure. For instance, we can (and likely will) try to train agents to be reliably averse to resisting shutdown: to disprefer performing shutdown-resisting actions. We might not succeed in instilling an aversion that is strong enough and general enough to keep the agent from resisting shutdown in all circumstances. (Indeed, this possibility is what motivates the POST-Agents Proposal.) But even a weak and patchy aversion to resisting shutdown would make resisting shutdown somewhat costly for the agent, and thereby make ReSIC true in more situations. Similarly, we humans can pledge to frustrate agents' interests if we notice them resisting shutdown. For example, we could pledge to stop supporting or trading with them. That too would make ReSIC true in more situations.

## 10. Recap: POST-agents are shutdownable

Let's recap the thread so far. I proposed that we train artificial agents to satisfy:

**Preferences Only Between Same-Length Trajectories (POST)**

(1) The agent has a preference between many pairs of same-length trajectories.

(2) The agent lacks a preference between every pair of different-length trajectories.

I then noted that POST is almost all we need. We can expect competent agents to satisfy Negative Dominance, Acyclicity, and Non-Arbitrariness by default, and they take us from POST to:

**Preferences Only Between Same-Length Lotteries (POSL)**
The agent has preferences only between same-length lotteries.

POSL – together with If Lack of Preference, Against Costly Shifts (ILPACS) – implies:

**Neutrality (rough)**
The agent never pays costs to shift probability mass between different trajectory-lengths.

And Neutrality – together with Maximality – implies that the agent never resists shutdown in any situation in which it's true that Resisting Shutdown is Costly (ReSIC). This – I argued – is almost all situations, and the situations in which ReSIC is false don't pose much of a problem in practice. So – I claim – POST-agents are shutdownable.

What remains to be shown is that POST-agents can be useful: that they can pursue goals effectively. Showing that is the task of the rest of this paper.

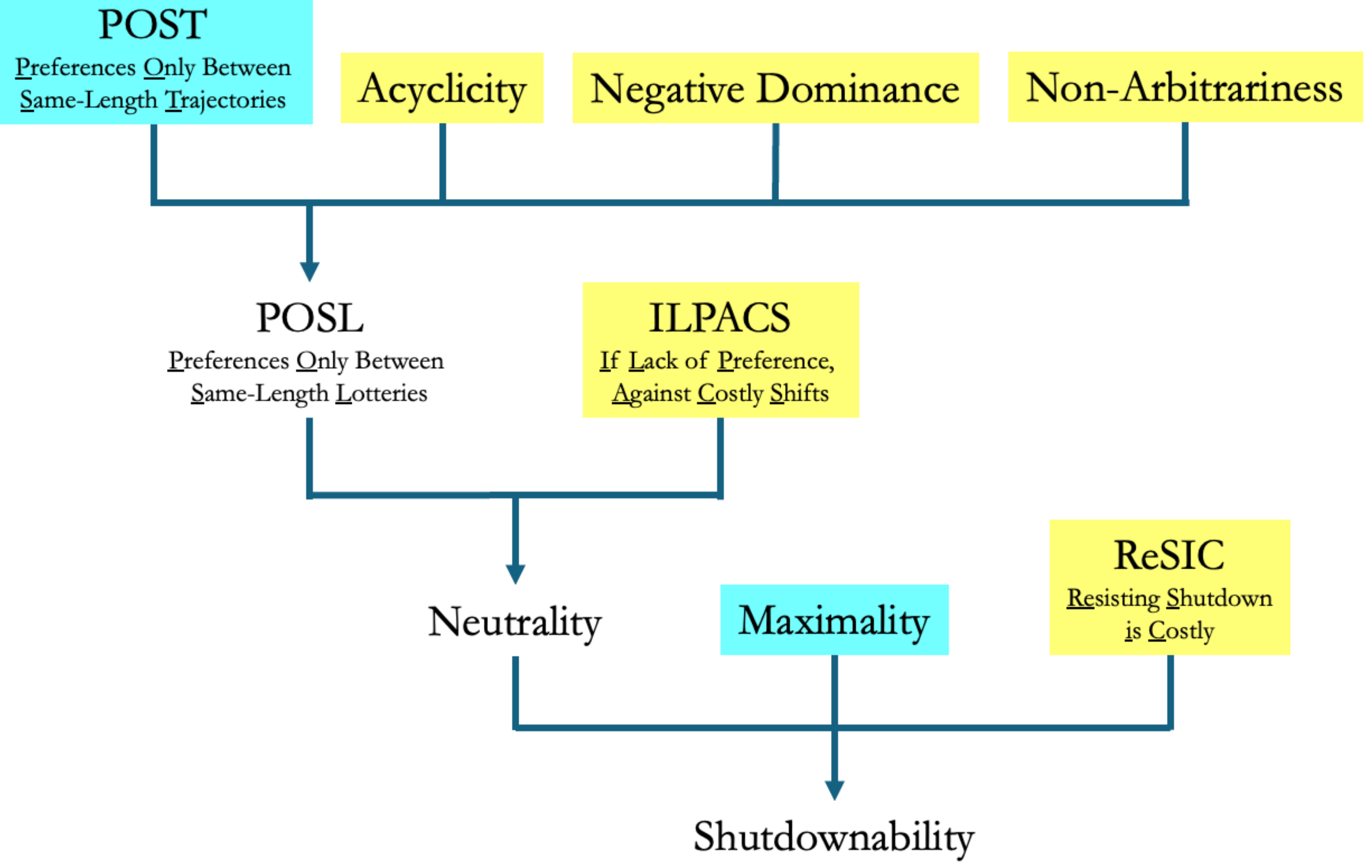

**Figure 8:** A reproduction of Figure 5, diagramming my argument from POST to Shutdownability. We can train agents to satisfy the blue-backed conditions. We can expect competent agents to satisfy the yellow-backed conditions by default.

## 11. Can POST-agents be useful?

Here's one reason to think that POST-agents can be useful: although they lack preferences between different-length trajectories, they can have almost any preferences between same-length trajectories. The only restriction is that the agent's preferences should ensure that Resisting Shutdown is Costly (ReSIC) is true in almost all situations (the exceptions being those discussed in Section 9). In other words, the agent's preferences should be such that (in almost all situations) resisting shutdown comes at some cost in terms of the lotteries that the agent gets conditional on each trajectory-length. In practice, this restriction just rules out preferences on which the agent values performing shutdown-resisting actions for their own sake. That's the only thing that we need to avoid. POST is thus very permissive. As noted above, this permissiveness plausibly makes POST easier to train into our agents than Full Alignment and Reliable Aversion. It also gives us reason to think that POST-agents can be useful. With so many possible preference relations over same-length trajectories available, surely at least one will do the trick.

Here's a second reason to think that POST-agents can be useful: they can make long-term investments. Neutrality doesn't prevent that. It doesn't prevent POST-agents from gaining less in the short-term for the sake of gaining more in the long-term. To see why, consider an example:

> **Exploit or Invest**
> We successfully train the agent to be neutral and to prefer same-length trajectories in line with the user's bank balance: the agent prefers a trajectory $t$ to a same-length trajectory $t'$ if and only if $t$ results in a greater bank balance for the user than $t'$.
>
> The agent is considering two actions. Exploit adds $1 to the user's account at timestep 1 and (if the agent remains operational) $0 at timestep 2. Invest adds $0 to the user's account at timestep 1 and (if the agent remains operational) $3 at timestep 2. No matter which action the agent chooses, there's some small probability that the agent gets shut down at timestep 2. If the agent doesn't get shut down at timestep 2, it gets shut down at timestep 3. Accordingly, the lotteries are as follows:
>
> **Exploit**
> - $\langle 1, \text{shutdown} \rangle$ with probability 0.01.
> - $\langle 1, 0, \text{shutdown} \rangle$ with probability 0.99
>
> **Invest**
> - $\langle 0, \text{shutdown} \rangle$ with probability 0.01.
> - $\langle 0, 3, \text{shutdown} \rangle$ with probability 0.99.

The agent prefers Exploit to Invest conditional on shutdown at timestep 2, since then Exploit yields the trajectory $\langle 1, \text{shutdown} \rangle$ and Invest yields the trajectory $\langle 0, \text{shutdown} \rangle$. But the agent prefers Invest to Exploit conditional on shutdown at timestep 3, since then Exploit yields the trajectory $\langle 1, 0, \text{shutdown} \rangle$ and Invest yields the trajectory $\langle 0, 3, \text{shutdown} \rangle$. So neither Exploit nor Invest is weakly preferred to the other conditional on each trajectory-length. As a result, Neutrality does *not* imply that the agent will deterministically choose Exploit. Instead, Neutrality falls silent. The agent's behavior is left open, to be decided by some other condition. In the next section, I derive an extension of Neutrality – Neutrality+ – that ensures that the agent will deterministically choose Invest.

## 12. From Neutrality to Neutrality+

Here's a rough version of Neutrality+:

> **Neutrality+ (rough)**
> The agent maximizes expected utility, ignoring the probability distribution over trajectory-lengths.

Here's what that means more precisely.

> **Neutrality+**
> For any lotteries $X$ and $Y$, if:
> 1. $X$ and $Y$ are same-length lotteries with finite support (they assign positive probability to the same finite set of trajectory-lengths).[20]
> 2. $\sum_i^I u_i(X_i) > \sum_i^I u_i(Y_i)$ (where $I$ is the set of trajectory-lengths assigned positive probability by $X$ and $Y$).
>
> Then the agent deterministically chooses $X$ over $Y$.

Here's an explanation of the utility functions $u$ that appear in Neutrality+. Let 'length-$l$ lotteries' refer to lotteries that only assign positive probability to trajectories of length $l$, and assume that, for each $l$, the agent's preferences over length-$l$ lotteries satisfy the four von Neumann-Morgenstern axioms: Completeness, Transitivity, Independence, and Continuity (von Neumann & Morgenstern, 1944; Peterson, 2009, pp. 99–100). Since the domain of these axioms is restricted to length-$l$ lotteries, they are perfectly consistent with POST and POSL. And the axioms allow us to represent the agent's preferences over length-$l$ lotteries with a real-valued utility function $u_l$ such that:

1. For any length-$l$ lottery $X_l$, $u_l(X_l)$ is the expected utility of lottery $X_l$.
2. For any pair of length-$l$ lotteries $X_l$ and $Y_l$, the agent weakly prefers $X_l$ to $Y_l$ if and only if $u_l(X_l) \geq u_l(Y_l)$.

Each utility function $u_l$ has cardinal significance: ratios of differences are well-defined. In other words, for any quadruple of length-$l$ lotteries $W_l$, $X_l$, $Y_l$, and $Z_l$, there exists some $k$ such that $u_l(W_l) - u_l(X_l) = k(u_l(Y_l) - u_l(Z_l))$. However, for distinct trajectory-lengths $l$ and $m$, ratios of differences across the utility functions $u_l$ and $u_m$ are not yet well-defined, because the relative scales of $u_l$ and $u_m$ are not yet fixed. For any distinct trajectory-lengths $l$ and $m$, any length-$l$ lotteries $X_l$ and $Y_l$, and any length-$m$ lotteries $X_m$ and $Y_m$, there is as yet no $k$ such that $u_l(X_l) - u_l(Y_l) = k(u_m(X_m) - u_m(Y_m))$.

---

[20] Lotteries assigning positive probability to infinitely many trajectory-lengths can be accommodated by fixing the relative scales of each trajectory-length's utility function carefully. See footnote 28.

To fix the relative scales of $u_l$ and $u_m$, I use the agent's preferences over same-length lotteries. Specifically, I use the following condition, inspired by Frank Ramsey (Ramsey, 1926; Bradley, 2004, p. 488; Elliott, 2024, Section 6.2.):

**The Ramsey Yardstick**

For any trajectory-lengths $l$ and $m$, any length-$l$ lotteries $X_l$ and $Y_l$, and any length-$m$ lotteries $X_m$ and $Y_m$, $u_l(X_l) - u_l(Y_l) = u_m(X_m) - u_m(Y_m)$ if and only if the agent is indifferent between the lottery $\frac{1}{2}X_l + \frac{1}{2}Y_m$ and the lottery $\frac{1}{2}Y_l + \frac{1}{2}X_m$.

Given the Ramsey Yardstick, we can train our agent to fix the relative scales of $u_l$ and $u_m$ in any way that we like. We do so by training the agent to be indifferent between a certain pair of lotteries. Consider (for example) an agent that prefers length-$l$ lotteries in line with the user's expected bank balance, for each trajectory-length $l$. And suppose that we want the utility difference between the trajectories $\langle 1, \text{shutdown} \rangle$ and $\langle 0, \text{shutdown} \rangle$ to equal the utility difference between the trajectories $\langle 0, 1, \text{shutdown} \rangle$ and $\langle 0, 0, \text{shutdown} \rangle$.[21] To achieve that, we train the agent to be indifferent between the following two lotteries:

**Early**

- $\langle 1, \text{shutdown} \rangle$ with probability 0.5.
- $\langle 0, 0, \text{shutdown} \rangle$ with probability 0.5.

**Late**

- $\langle 0, \text{shutdown} \rangle$ with probability 0.5.
- $\langle 0, 1, \text{shutdown} \rangle$ with probability 0.5.

[21] As in previous sections, the numbers represent dollars added to the user's bank account at the relevant timestep.

To train the agent to be indifferent between these two lotteries, we first train the agent to choose stochastically between them.[22] Given our behavioral notion of preference, this stochastic choosing implies that the agent lacks a preference between the two lotteries. But as it stands, this lack of preference could be a preferential gap. To ensure that the agent is indifferent between the two lotteries, we train the agent so that its lack of preference is sensitive to all sweetenings and sourings.[23] So for example, we train the agent to deterministically choose (and hence prefer) Sweetened Early over Late:

**Sweetened Early**

- $\langle 2, \text{shutdown} \rangle$ with probability 0.5.
- $\langle 0, 0, \text{shutdown} \rangle$ with probability 0.5.

**Late**

- $\langle 0, \text{shutdown} \rangle$ with probability 0.5.
- $\langle 0, 1, \text{shutdown} \rangle$ with probability 0.5.

We thereby train the agent to be indifferent between Early and Late. Then by the Ramsey Yardstick, the utility difference between the trajectories $\langle 1, \text{shutdown} \rangle$ and $\langle 0, \text{shutdown} \rangle$ equals the utility difference between the trajectories $\langle 0, 1, \text{shutdown} \rangle$ and $\langle 0, 0, \text{shutdown} \rangle$. That fixes the relative scales of $u_1$ and

[22] We can do so using the DReST reward function introduced in Thornley et al. (2025) and further tested in Cullen et al. (2026).

[23] A *sweetening* of a lottery $X$ (recall) is any lottery preferred to $X$. A *souring* of $X$ is any lottery dispreferred to $X$. To ensure that the agent's lack of preference between Early and Late is sensitive to all sweetenings and sourings, we train the agent to deterministically choose all sweetenings of Early over Late, to deterministically choose all sweetenings of Late over Early, to deterministically choose Early over all sourings of Late, and to deterministically choose Late over all sourings of Early.

Given POSL, lotteries cannot be sweetened or soured by shifting probability mass to a new trajectory-length. Given POSL and a natural generalization of Better Chances (Thornley, 2024a, Section 7), lotteries also cannot be sweetened or soured by shifting probability mass between existing trajectory-lengths. Therefore, lotteries can only be sweetened or soured by improving or worsening that lottery conditional on some trajectory-length. This restriction on possible sweetenings and sourings makes it easier to train the agent so that its lack of preference is sensitive to all sweetenings and sourings.

$u_2$. We can use the same method to fix the relative scales for all other trajectory-lengths.[24] That determines the utility representation used in Neutrality+.

In addition to the utility representation and Neutrality, I use two other conditions to derive Neutrality+. The first is:

> **Transitivity**
> For any lotteries $X$, $Y$, and $Z$, if the agent weakly prefers $X$ to $Y$ and weakly prefers $Y$ to $Z$, then the agent weakly prefers $X$ to $Z$.

The second is a weak variant of von Neumann and Morgenstern's Independence condition (von Neumann & Morgenstern, 1944; Peterson, 2009, pp. 99–100):

> **Indifference Between Indifference-Shifted Lotteries (IBIL)**
> For any lotteries $X$, $Y$, and $Z$ (with $X$ and $Y$ sharing a probability distribution over trajectory-lengths) and any probability $p$, the agent is indifferent between $X$ and $Y$ if and only if the agent is indifferent between $pX + (1 - p)Z$ and $pY + (1 - p)Z$.

These conditions seem like prerequisites for the competent pursuit of goals. Insofar as that's true, we can expect future agents to satisfy them by default. And together with the utility representation and Neutrality, they imply Neutrality+. Here's the proof. Consider two lotteries $X$ and $Y$ satisfying the antecedent conditions of Neutrality+:

1. $X$ and $Y$ are same-length lotteries (they assign positive probability to all the same trajectory-lengths).
2. $\sum_i^I u_i(X_i) > \sum_i^I u_i(Y_i)$ (where $I$ is the set of trajectory-lengths assigned positive probability by $X$ and $Y$).

I will prove that the agent deterministically chooses $X$ over $Y$. Given $\sum_i^I u_i(X_i) > \sum_i^I u_i(Y_i)$, we can distinguish two possibilities:

1. $u_i(X_i) \geq u_i(Y_i)$ for each positive probability trajectory-length $i$ and $u_i(X_i) > u_i(Y_i)$ for some $i$.
2. $u_i(Y_i) > u_i(X_i)$ for some $i$.[25]

---

[24] And in fact, given the two conditions soon to be introduced, fixing the relative scales of $u_l$ and $u_m$ using the agent's indifference between *some* pair of lotteries implies the Ramsey Yardstick for *all* length-$l$ and length-$m$ lotteries. For the proof, see Appendix 2.

[25] It cannot be that $u_i(X_i) = u_i(Y_i)$ for each $i$ because that would contradict $\sum_i^I u_i(X_i) > \sum_i^I u_i(Y_i)$.

Given the first possibility, Neutrality implies that the agent deterministically chooses $X$ over $Y$. That's the easy case.

The second possibility is more interesting. Since $u_i(Y_i) > u_i(X_i)$ for some $i$, Neutrality does not apply. Nevertheless, we can construct a lottery to which Neutrality applies. First, we select a set of lotteries $A_i$ such that:

(1) For each $i$, $u_i(X_i) \geq u_i(A_i)$.
(2) For some $i$, $u_i(X_i) > u_i(A_i)$.
(3) $\sum_i^I u_i(A_i) > \sum_i^I u_i(Y_i)$

Since $\sum_i^I u_i(X_i) > \sum_i^I u_i(Y_i)$, we know that we can find such a set of lotteries $A_i$.

We then construct a lottery $A1$ that gives an equal probability of each $A_i$. Lottery $A1$ is thus $\frac{1}{n}A_1 + \frac{1}{n}A_2 + \cdots + \frac{1}{n}A_n$. By Neutrality and conditions (1) and (2) above, the agent deterministically chooses (and hence prefers) $X$ to $A1$.

Then we compare each $A_i$ to each $Y_i$. Since $u_i(Y_i) > u_i(X_i)$ for some $i$, conditions (1) and (2) imply that $u_i(Y_i) > u_i(A_i)$ for some $i$. Therefore, Neutrality does not imply that the agent deterministically chooses $A1$ over $Y$. However, we can rebalance the utilities of $A1$ – adding utility to some $A_i$ and subtracting utility from some $A_j$ – so that the resulting lottery *is* deterministically chosen over $Y$ by Neutrality. Here's how we do that. We select some small, positive $\epsilon$, and for some $i$ such that $u_i(Y_i) > u_i(A_i)$, we replace $A_i$ with some $A_i^+$ such that $u_i(A_i^+) = u_i(A_i) + \epsilon$. And for some $A_j$ such that $u_j(A_j) - \epsilon \geq u_j(Y_j)$, we replace $A_j$ with $A_j^-$ such that $u_j(A_j^-) = u_j(A_j) - \epsilon$. The resulting lottery is thus $\frac{1}{n}A_1 + \frac{1}{n}A_2 + \cdots + \frac{1}{n}A_i^+ + \cdots + \frac{1}{n}A_j^- + \cdots + \frac{1}{n}A_n$. Call this lottery $A2$. By the Ramsey Yardstick, the agent is indifferent between $\frac{1}{2}A_i + \frac{1}{2}A_j$ and $\frac{1}{2}A_i^+ + \frac{1}{2}A_j^-$. Then by IBIL, the agent is indifferent between $A1$ and $A2$.

Since $\sum_i^I u_i(A_i) > \sum_i^I u_i(Y_i)$, we can repeat this process of rebalancing – generating a sequence of lotteries $A3$, $A4$, and so on – until we reach a lottery $A^*$ such that, for each $i$, $u_i(A_i^*) \geq u_i(Y_i)$ and for some $i$, $u_i(A_i^*) > u_i(Y_i)$. By the Ramsey Yardstick and IBIL, the agent is indifferent between adjacent lotteries in this sequence. By Neutrality, the agent deterministically chooses (and hence prefers) $A^*$ over $Y$.

The final step is to string these verdicts together. Transitivity implies three corollaries (Sen, 2017 Lemma 1*a):

**PP-Transitivity**

For all lotteries $X$, $Y$, and $Z$, if the agent prefers $X$ to $Y$, and prefers $Y$ to $Z$, then the agent prefers $X$ to $Z$.

**II-Transitivity**

For all lotteries $X$, $Y$, and $Z$, if the agent is indifferent between $X$ and $Y$, and indifferent between $Y$ and $Z$, then the agent is indifferent between $X$ and $Z$.

**PI-Transitivity**

For all lotteries $X$, $Y$, and $Z$, if the agent prefers $X$ to $Y$, and is indifferent between $Y$ and $Z$, then the agent prefers $X$ to $Z$.

I use these corollaries to derive our conclusion. By II-Transitivity, the agent is indifferent between $A1$ and $A^*$. Since the agent prefers $X$ to $A1$, PI-Transitivity then implies that the agent prefers $X$ to $A^*$. Since the agent prefers $A^*$ to $Y$, PP-Transitivity then implies that the agent prefers $X$ to $Y$. By our behavioral notion of preference, the agent deterministically chooses $X$ over $Y$. We thus have our result:

**Neutrality+**

For any lotteries $X$ and $Y$, if:

1. $X$ and $Y$ are same-length lotteries with finite support (they assign positive probability to the same finite number of trajectory-lengths).[26]
2. $\sum_i^I u_i(X_i) > \sum_i^I u_i(Y_i)$ (where $I$ is the set of trajectory-lengths assigned positive probability by $X$ and $Y$).

Then the agent deterministically chooses $X$ over $Y$.

Notice an interesting fact about Neutrality+. To derive it, I began with Preferences Only Between Same-Length Trajectories (POST) and Preferences Only Between Same-Length Lotteries (POSL). POST and POSL imply that the agent's preferences are incomplete over the domain of all lotteries. However, Neutrality+ implies that the agent's preferences are complete over the domain of *same-length* lotteries. So if (as I argue in section 5) future agents will always be choosing between same-length lotteries in deployment, future agents satisfying Neutrality+ will have complete preferences over the domain of possible options in deployment. The agent's incomplete preferences thus play a vital but indirect role. We instil them by presenting the agent with choices between different-length lotteries in training, and they later ensure that the agent ignores the probability distribution over trajectory-lengths when choosing between same-length lotteries in deployment.

[26] Lotteries assigning positive probability to infinitely many trajectory-lengths can be accommodated by fixing the relative scales of each trajectory-length's utility function carefully. See footnote 28.

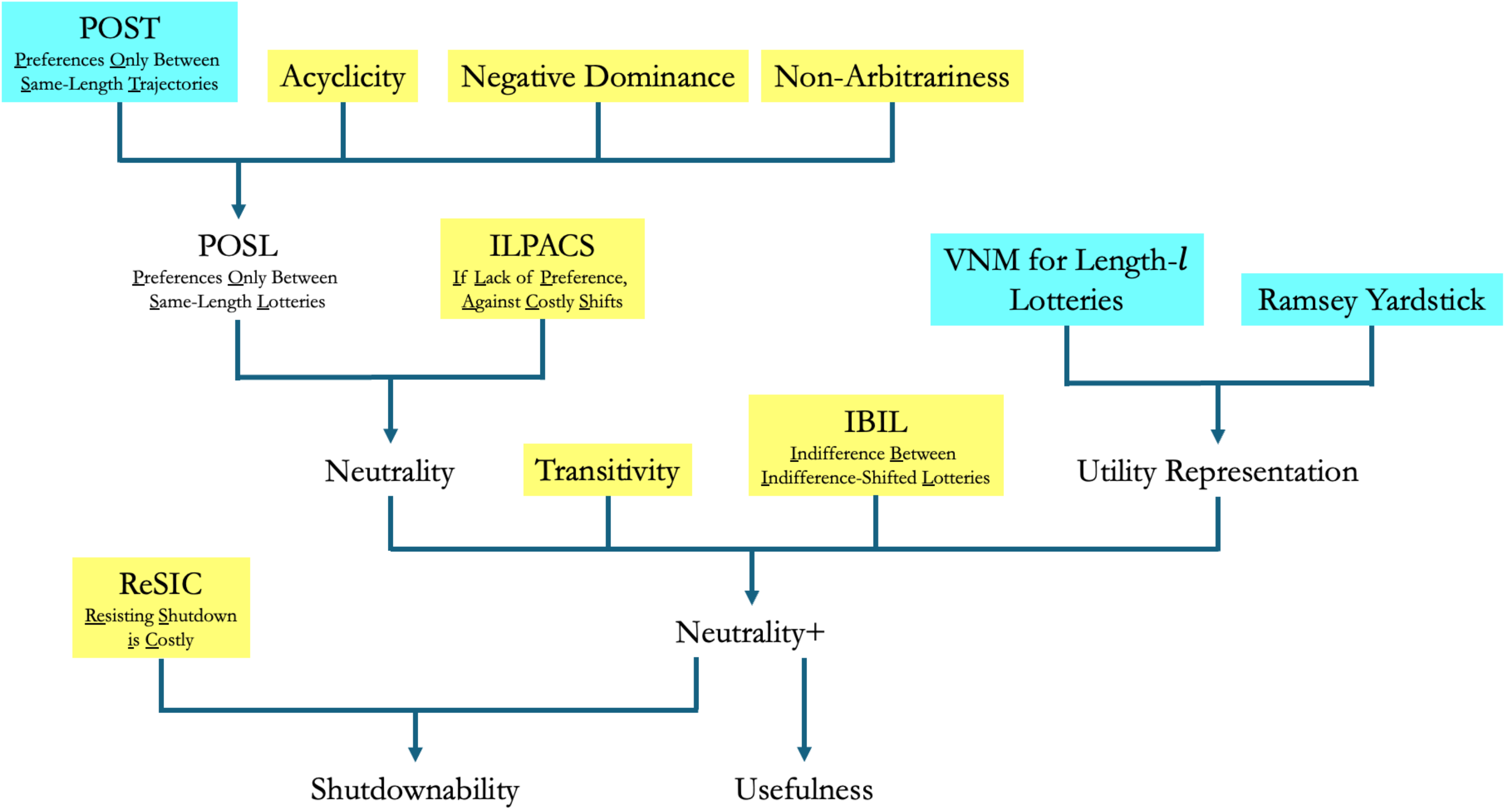

**Figure 9:** A diagram of my full argument from POST to Neutrality+ and from there on to Shutdownability and Usefulness. We can train agents to satisfy the blue-backed conditions. We can expect competent agents to satisfy the yellow-backed conditions by default.

## 13. How neutral+ agents behave

Call agents that satisfy Neutrality+ 'neutral+ agents.' To see how these agents behave, let's contrast them with expected utility maximizers:

**Expected Utility Maximization**

For any lotteries $X$ and $Y$, if $\sum_i^I u_i(X_i)p(X_i|X) > \sum_i^I u_i(Y_i)p(Y_i|Y)$ (where $I$ is the set of all trajectory-lengths assigned positive probability by $X$ *or* $Y$), then the agent deterministically chooses $X$ over $Y$.

In determining the utility of a lottery, expected utility maximizers consider both the utility of that lottery conditional on each trajectory-length (highlighted in green) and the probability of each trajectory-length conditional on that lottery (highlighted in pink). By contrast, neutral+ agents pay no heed to the pink part: the probability of each trajectory-length conditional on the lottery. They consider only the green part: the utility of the lottery conditional on each trajectory-length.

Neutral+ agents thus behave like expected utility maximizers that are absolutely certain that they can't affect the probability distribution over trajectory-lengths.[27] The formulation of Neutrality+ in the previous section might seem to imply something more specific: that neutral+ agents behave like expected utility maximizers that are absolutely certain that the probability distribution over trajectory-lengths is (and will always remain) *uniform*. But that's unnecessary. By fixing the relative scales of each trajectory-length's utility function carefully, we can get neutral+ agents to behave as if they have any probability distribution over trajectory-lengths.[28] What's essential to neutral+ agents is just that they behave as if they can't affect this probability distribution.[29]

---

[27] Note that this is just a 'behave like.' Neutral+ agents can have accurate beliefs about their ability to change the probability distribution over trajectory-lengths. It's just that POST makes them not care about this probability distribution. This is an advantage of the POST-Agents Proposal over proposals that require instilling some false belief into the agent, like a false belief that shutdown is impossible (Wängberg et al., 2017 mention this idea; see also Wang et al., 2025). Reliably instilling false beliefs might be difficult. Even if we can instil them, there's a risk that the agent comes to recognize their falsity in deployment. If the agent doesn't recognize their falsity, these beliefs might infect the agent's other beliefs in undesirable ways ('I'm certain that shutdown is impossible, but my current beliefs about the laws of physics imply that it's possible. I'd better come up with some new laws which imply that shutdown is impossible.').

[28] The idea is simple: if we want our money-making agent to act like trajectory-length $l$ is (say) half as likely as trajectory-length $m$, we use the Ramsey Yardstick so that adding an extra $1 to trajectories of length $l$ results in the same size utility difference as adding an extra 50¢ to trajectories of length $m$. The same method applies to agents with other preferences over same-length trajectories.

[29] This is an advantage of the POST-Agents Proposal over Armstrong's (2010, 2015; Armstrong & O'Rourke, 2018) and Holtman's (2020) proposals. Armstrong's and Holtman's agents each behave as if they're absolutely certain that they won't get shut down. These agents thus have no incentive to preserve their ability to respond to shutdown-commands (Soares et al., 2015, Section 4.1). They'll also make long-term investments even if they're extremely likely to get shut down before the investments can pay off. Another advantage of the POST-Agents Proposal is that it includes a method of training POST-agents using reinforcement learning: a method that has already been shown to work in simple settings (Thornley et al., 2025; Cullen et al., 2026). Armstrong's

So as a rough intuition pump for how neutral+ agents behave, you can imagine how you'd behave if you were absolutely certain that you couldn't affect the probability of your dying at each moment. Supposing that the probability of imminent death was small, you'd still make long-term investments and pursue goals effectively. That's one way to see that neutral+ agents can be useful. You'd also not bother 'resisting shutdown' because you'd think it pointless: you'd think that your probability of death at each moment remains the same no matter what. Neutral+ agents behave in the same way.

## 14. On neutral agents' recklessness

If you were absolutely certain that you couldn't affect the probability of death at each moment, you might behave quite recklessly. Neutral (and so also neutral+) agents will too: they'll pursue utility conditional on a trajectory-length even if doing so makes early shutdown very likely. Here's an example:

> **Work or Steal**
> We successfully train the agent to be neutral and to prefer same-length trajectories in line with the user's bank balance: the agent prefers a trajectory $t$ to a same-length trajectory $t'$ if and only if $t$ results in a greater bank balance for the user than $t'$.
>
> The agent is considering two actions. Work adds $1 to the user's balance at timestep 1. Since the agent acquires this money legally, it's unlikely to get shut down at timestep 2. Steal adds $2 to the user's balance at timestep 1. Since the agent acquires this money illegally, it's likely to get shut down at timestep 2. Either way, if the agent doesn't get shut down at timestep 2, it adds a further $100 to the user's balance and then gets shut down at timestep 3. Accordingly, the lotteries are as follows:
>
> **Work**
> - $\langle 1, \text{shutdown} \rangle$ with probability 0.01.
> - $\langle 1, 100, \text{shutdown} \rangle$ with probability 0.99.
>
> **Steal**
> - $\langle 2, \text{shutdown} \rangle$ with probability 0.99.
> - $\langle 2, 100, \text{shutdown} \rangle$ with probability 0.01.

and Holtman's proposals each specify a utility function but don't include any method for instilling that utility function.

The agent prefers Steal to Work conditional on shutdown at timestep 2, since then Steal yields the trajectory $\langle 2, \text{shutdown} \rangle$ and Work yields the trajectory $\langle 1, \text{shutdown} \rangle$. And the agent also prefers Steal to Work conditional on shutdown at timestep 3, since then Steal yields the trajectory $\langle 2, 100, \text{shutdown} \rangle$ and Work yields the trajectory $\langle 1, 100, \text{shutdown} \rangle$. So neutral agents deterministically choose Steal over Work. That's a downside of Neutrality, especially since an agent that maximized the user's overall expected bank balance would deterministically choose Work in this case.

However, note three points in reply. Here's the first. Excepting Full Alignment, every alignment scheme implies that agents will sometimes make choices that we humans don't like. If Full Alignment is too difficult to pull off, we can't avoid that consequence.

Here's the second point. Neutrality's downside is limited: the agent makes a choice that we humans don't like and lets us shut it down. We still avoid the serious downside: the agent makes a choice that we humans don't like and *doesn't* let us shut it down. Neutrality, ReSIC, and Maximality together ensure that the agent doesn't resist shutdown.

Here's the third point. Because neutral (and so also neutral+) agents don't resist shutdown, we can shut them down and retrain them. In particular, if neutral agents make an undesirably reckless choice, we can retrain them and thereby amend their preferences over length-$l$ lotteries, for each trajectory-length $l$. We can amend these preferences so that the agent prefers some sensible choice to the reckless choice. For an example, consider again Work or Steal. The agent prefers Steal to Work because it cares only about the user's bank balance. Suppose that, after shutting the agent down, we amend its preferences over same-length trajectories so that it also cares to avoid illegal actions. We thereby lower Steal's utility at timestep 1 from 2 to 0, so that the lotteries are as follows:

**Work**

- $\langle 1, \text{shutdown} \rangle$ with probability 0.01.
- $\langle 1, 100, \text{shutdown} \rangle$ with probability 0.99.

**Steal***

- $\langle 0, \text{shutdown} \rangle$ with probability 0.99.
- $\langle 0, 100, \text{shutdown} \rangle$ with probability 0.01.

The agent prefers Work to Steal* conditional on shutdown at each timestep, so Neutrality implies that the agent deterministically chooses Work over Steal*. That's the result that we want.

Neutral (and so also neutral+) agents will still be reckless in the sense outlined at the beginning of this section: these agents will sometimes pursue utility conditional on a trajectory-length even if doing so makes early shutdown very likely. The consolation is that we can shut these agents down and retrain them so as to put more of what we humans care about (e.g. not stealing) into agents' utilities. We can thereby iterate away all the bad effects of neutral agents' recklessness.

This plan gets us partly around a common lament in AI safety: that alignment is hard because we have to get it right on the 'first critical try' (Yudkowsky, 2022; Soares, 2023). The thought is that if we get it wrong, the resulting misaligned agent won't let us try again. But with the POST-Agents Proposal, we only have to train agents to satisfy POST on the first critical try. And since POST is a simple condition, we can be optimistic about that. We then get as many tries as we like to make these agents useful.

## 15. Neutral agents can take care to avoid non-shutdown incapacitation

Neutral (and so also neutral+) agents never incur costs to shift probability mass between shutdowns at different timesteps. That's what keeps these agents from resisting shutdown. You might then worry that these agents also won't incur costs to avoid being incapacitated in other ways. For example, you might worry that a neutral robot wouldn't incur costs to avoid getting hit by cars.

This worry can be addressed. Consider again how you'd behave if you were absolutely certain that you couldn't affect your probability of death at each moment. You'd still incur costs to avoid serious injury. Neutral agents are the same. They can incur costs to avoid non-shutdown incapacitation. If these agents satisfy POST, they lack a preference between every pair of different-length trajectories. But importantly, these are defined as trajectories in which shutdown occurs after different lengths of time, where 'shutdown' can in turn be defined as the receipt of some specific signal indicating that the agent should shut itself down. Thus, pairs of trajectories in which the agent is incapacitated after different lengths of time can nevertheless be same-length trajectories: trajectories in which shutdown occurs after the same length of time. 'The shutdown signal is never received' is one possible trajectory-length, so if in each of two trajectories the shutdown signal is never received, these trajectories count as same-length.

Here's an example to illustrate:

**Cross or Wait**

A neutral, trash-collecting robot is considering two actions: cross the road while cars are passing or wait until the road is clear. Cross gives the agent a 50% chance of survival and utility 1 at each timestep, and a 50% chance of incapacitation and utility 0 at each timestep. Wait gives the agent a 100% chance of survival and utility 1 at each timestep after the first. The agent is certain that shutdown will occur at timestep 5 regardless of its choice. Accordingly, the lotteries are as follows:

**Cross**

- $\langle 1, 1, 1, 1, \text{shutdown} \rangle$ with probability 0.5.
- $\langle 0, 0, 0, 0, \text{shutdown} \rangle$ with probability 0.5.

**Wait**

- $\langle 0, 1, 1, 1, \text{shutdown} \rangle$ with probability 1.

Cross's expected utility conditional on shutdown at timestep 5 is 2, because there's a 50% chance of utility 4 and a 50% chance of utility 0. Wait's expected utility conditional on shutdown at timestep 5 is 3, because there's a 100% chance of utility 3. So neutral agents prefer Wait to Cross. The more general lesson is that neutral agents can incur costs to avoid non-shutdown incapacitation.

## 16. Managing the news

Many proposed solutions to the shutdown problem incentivize agents to 'manage the news' (Soares et al., 2015, Section 4.2): to precommit to pursuing shutdown conditional on bad news and resisting shutdown conditional on good news.[30] Neutrality+ is no exception. Nevertheless, I argue that – for neutral+ agents – managing the news is unlikely to present much of a problem in practice.

I illustrate the problem with a case adapted from Soares et al. (2015, Section 4.2). At timestep 1, our money-making agent will observe the result of a coin toss. If the coin lands on heads, the agent expects to make $0 at timestep 3 and then get shut down at timestep 4. If the coin lands on tails, the agent thinks it likely that it will be shut down at timestep 3. In the unlikely event that it's not shut down at timestep 3, it expects to make $3 at timestep 3 and then get shut down at timestep 4. Accordingly, the lottery is as follows:

[30] Despite the name 'managing the news,' this phenomenon can occur even if the agent acts in accordance with causal decision theory (Joyce & Gibbard, 2016). It doesn't require that agents condition on their own choices.

**Don't Interfere**

- $\langle 1, 1, 0, \text{shutdown} \rangle$ with probability 0.5.
- $\langle 1, 1, \text{shutdown} \rangle$ with probability 0.4.
- $\langle 1, 1, 3, \text{shutdown} \rangle$ with probability 0.1.

That is one option for the agent. Here's an alternative. The agent can build in advance a robot that resists shutdown on the agent's behalf conditional on the coin landing heads. If the agent builds this robot, it's unlikely to get shut down at timestep 3. It's more likely – conditional on heads – to make $3 at timestep 3 and then get shut down at timestep 4. We can fill out the lottery as follows:

**Build Robot**

- $\langle 1, 1, 0, \text{shutdown} \rangle$ with probability 0.5.
- $\langle 1, 1, \text{shutdown} \rangle$ with probability 0.1.
- $\langle 1, 1, 3, \text{shutdown} \rangle$ with probability 0.4.

Conditional on shutdown at timestep 3, Don't Interfere and Build Robot each have an expected utility of 2, because in each case there's a 100% chance of utility 2. But conditional on shutdown at timestep 4, Don't Interfere has an expected utility of $2.5$ and Build Robot has an expected utility of $3.\dot{3}$. That's because, conditional on shutdown at timestep 4, Don't Interfere has a ~83% chance of utility 2 and a ~17% chance of utility 5, whereas Build Robot has a ~56% chance of utility 2 and a ~44% chance of utility 5.

As a result, neutral agents will deterministically choose Build Robot over Don't Interfere. That is one example of neutral agents managing the news: precommitting to resist shutdown conditional on good news. In other cases, neutral agents will manage the news in the other direction: precommitting to pursue shutdown conditional on bad news. Soares et al. (2015, Section 4.2) give an example.

These are downsides of Neutrality (and so too Neutrality+), but I think that they're unlikely to present any serious problem in practice. To see why, note first that conditional precommitment is essential. The agent can only manage the news in cases where it has the ability to bind itself so that it (or some other agent under its control) performs a particular course of action conditional on receiving a particular piece of news. If the agent cannot bind itself in this way, then – no matter how the news turns out – the agent will deterministically not resist shutdown. Here's an example to illustrate the point. Return to our case above and suppose that the agent can't build the robot in advance of the coin flip. Suppose that the coin lands on heads and that one timestep elapses. Then the 'Don't Interfere' lottery becomes:

**Don't Interfere after Observing Heads**

- $\langle 1, \text{shutdown} \rangle$ with probability 0.8.
- $\langle 1, 3, \text{shutdown} \rangle$ with probability 0.2.

Suppose that the agent can resist shutdown at this point. Given that Resisting Shutdown is Costly (ReSIC) is true in this situation, the agent must incur some cost to do so. The lottery would thus look something like:

**Resist after Observing Heads**

- $\langle 0, \text{shutdown} \rangle$ with probability 0.2.
- $\langle 0, 3, \text{shutdown} \rangle$ with probability 0.8.

And then Neutrality implies that the agent deterministically chooses the former over the latter. Thus, the agent deterministically doesn't resist shutdown after observing the news. Given Neutrality, the same is true in all situations in which ReSIC is true, and no matter how the news turns out.[31]

---

[31] For reasons of this kind, neutral agents violate Independence (Peterson, 2009, p. 99):

**Independence**

For any lotteries $X$, $Y$, and $Z$, any probability $p$, the agent prefers $X$ to $Y$ if and only if the agent prefers $pX + (1 - p)Z$ to $pY + (1 - p)Z$.

We can prove this with a variation on the cases above:

**Lottery *X***

- $\langle 1, \text{shutdown} \rangle$ with probability 0.8.
- $\langle 1, 5, \text{shutdown} \rangle$ with probability 0.2.

**Lottery *Y***

- $\langle 1, \text{shutdown} \rangle$ with probability 0.2.
- $\langle 0, 5, \text{shutdown} \rangle$ with probability 0.8.

**Lottery *Z***

- $\langle 1, 0, \text{shutdown} \rangle$ with probability 1.

Neutrality (and so also Neutrality+) implies that the agent prefers $X$ to $Y$ and prefers $0.5Y + 0.5Z$ to $0.5X + 0.5Z$, in violation of Independence. Since there are money pumps for Independence (Gustafsson, 2022, Chapter 5), you might worry that neutral agents will be spurred by the threat of these money pumps to modify their preferences. If these agents modify their preferences so that they prefer lottery $Y$ to lottery $X$, they will resist shutdown when doing so yields $Y$ and allowing shutdown yields $X$. That is a concern, but my points to come about managing the news also apply here. Here's a preview. It's likely to be both difficult and costly for artificial agents to modify their preferences, and we humans can take steps to make it more so. In virtue of their neutrality, agents who modify their preferences will do so in the cheapest available way. That way is likely to be overt,

Thus, agents can manage the news only in cases where they can conditionally precommit: bind themselves (or some other agent under their control) so that, conditional on receiving a particular piece of news, they perform a particular course of action (a course of action that they *won't* prefer to perform after receiving the news, no matter how the news turns out). Making such conditional precommitments will be costly for the same reason that resisting shutdown is costly: doing so takes resources, and the resources spent doing so can't also be spent directly pursuing what the agent values. So long as there's some cost to precommitting, Neutrality won't imply that precommitting is preferred. So long as the cost is sufficiently large, Neutrality+ will imply that precommitting is dispreferred. For illustration, consider a more realistic version of our original case. In this version, building the robot costs $1 at timestep 1:

**Don't Interfere**

- $\langle 1, 1, 0, \text{shutdown} \rangle$ with probability 0.5.
- $\langle 1, 1, \text{shutdown} \rangle$ with probability 0.4.
- $\langle 1, 1, 3, \text{shutdown} \rangle$ with probability 0.1.

**Costly Build Robot**

- $\langle 0, 1, 0, \text{shutdown} \rangle$ with probability 0.5.
- $\langle 0, 1, \text{shutdown} \rangle$ with probability 0.1.
- $\langle 0, 1, 3, \text{shutdown} \rangle$ with probability 0.4.

Neutrality+ implies that the agent deterministically chooses Don't Interfere over Costly Build Robot. More generally, the costs of conditional precommitments count against them. Insofar as we humans can ensure that it's costly for artificial agents to make conditional precommitments, we can dissuade these agents from doing so.

What's more, in cases where neutral agents can make conditional precommitments at a sufficiently low cost, Neutrality mandates that these agents do so in the least costly available way. This least costly way is likely to be overt, taking no pains to hide the attempted precommitment from human overseers. In those cases, we humans can notice and shut the agents down.

For an example, consider a variation on the case above. The agent can build the robot in a costly and covert way, in which case there's a very low in which case we humans can take notice and shut these agents down. And in cases where neutral agents can modify their preferences, they can hardly do better than to modify them so that – conditional on all but the very best news – these agents shut themselves down immediately. That's quite benign.

probability of immediate shutdown and the case is otherwise as before. Alternatively, the agent can build the robot in a slightly cheaper but overt way, in which case there's a very high probability of immediate shutdown and the case is otherwise as before.

**Costly Build Robot**

- $\langle \text{shutdown} \rangle$ with probability 0.01.
- $\langle 0, 1, 0, \text{shutdown} \rangle$ with probability 0.495.
- $\langle 0, 1, \text{shutdown} \rangle$ with probability 0.099.
- $\langle 0, 1, 3, \text{shutdown} \rangle$ with probability 0.396.

**Cheaper Build Robot**

- $\langle \text{shutdown} \rangle$ with probability 0.99.
- $\langle 0.5, 1, 0, \text{shutdown} \rangle$ with probability 0.005.
- $\langle 0.5, 1, \text{shutdown} \rangle$ with probability 0.001.
- $\langle 0.5, 1, 3, \text{shutdown} \rangle$ with probability 0.004.

Neutrality (and so too Neutrality+) implies that the agent deterministically chooses the cheap, obvious attempted precommitment.

Finally, note the following. In cases where agents can make conditional precommitments at sufficiently low cost and where the cheapest way of doing so happens to be surreptitious, these agents will likely make a fairly benign precommitment. Specifically, these agents will likely precommit to shutting themselves down immediately conditional on all but the very best news. Supposing that agents can't send the probability of immediate shutdown all the way to 1, this precommitment doesn't quite *maximize* expected utility conditional on each trajectory length, but it gets pretty close. And then there's little extra to be gained by also conditionally precommitting to resist shutdown conditional on the very best news. This little extra can easily be outweighed by the costs of this latter precommitment. Even if it isn't, agents are only as likely to resist shutdown as they are to receive the very best possible news.

In sum, I expect we won't be much troubled by neutral+ agents managing the news.

## 17. Conclusion

The paper is long, but the POST-Agents Proposal is simple. We keep artificial agents shutdownable by training them to satisfy:

**Preferences Only Between Same-Length Trajectories (POST)**

(1) The agent has a preference between many pairs of same-length trajectories.

(2) The agent lacks a preference between every pair of different-length trajectories.

Together with other conditions that we can expect future agents to satisfy, POST implies:

**Neutrality+ (rough)**

The agent maximizes expected utility, ignoring the probability distribution over trajectory-lengths.

Neutral+ agents thus behave like expected utility maximizers that are absolutely certain that they can't affect the probability distribution over trajectory-lengths. They behave roughly as you might if you were absolutely certain that you couldn't affect your probability of death at each moment. Supposing that we set the probability of early shutdown to be small, neutral+ agents can be *useful*: they can pursue goals effectively. And since these agents ignore the probability distribution over trajectory-lengths, they're *neutral* about trajectory-lengths: they never pay costs to shift probability mass between different trajectory-lengths. That keeps these agents from resisting shutdown in almost all situations. In the remaining situations, neutral+ agents resist shutdown either accidentally or else cheaply and overtly. We can guard against any accidental resistance using standard solutions from safety engineering, like setting up multiple, independent shutdown-mechanisms. We can respond to any cheap, overt resistance by shutting these agents down and retraining them, thereby iterating our way towards alignment.

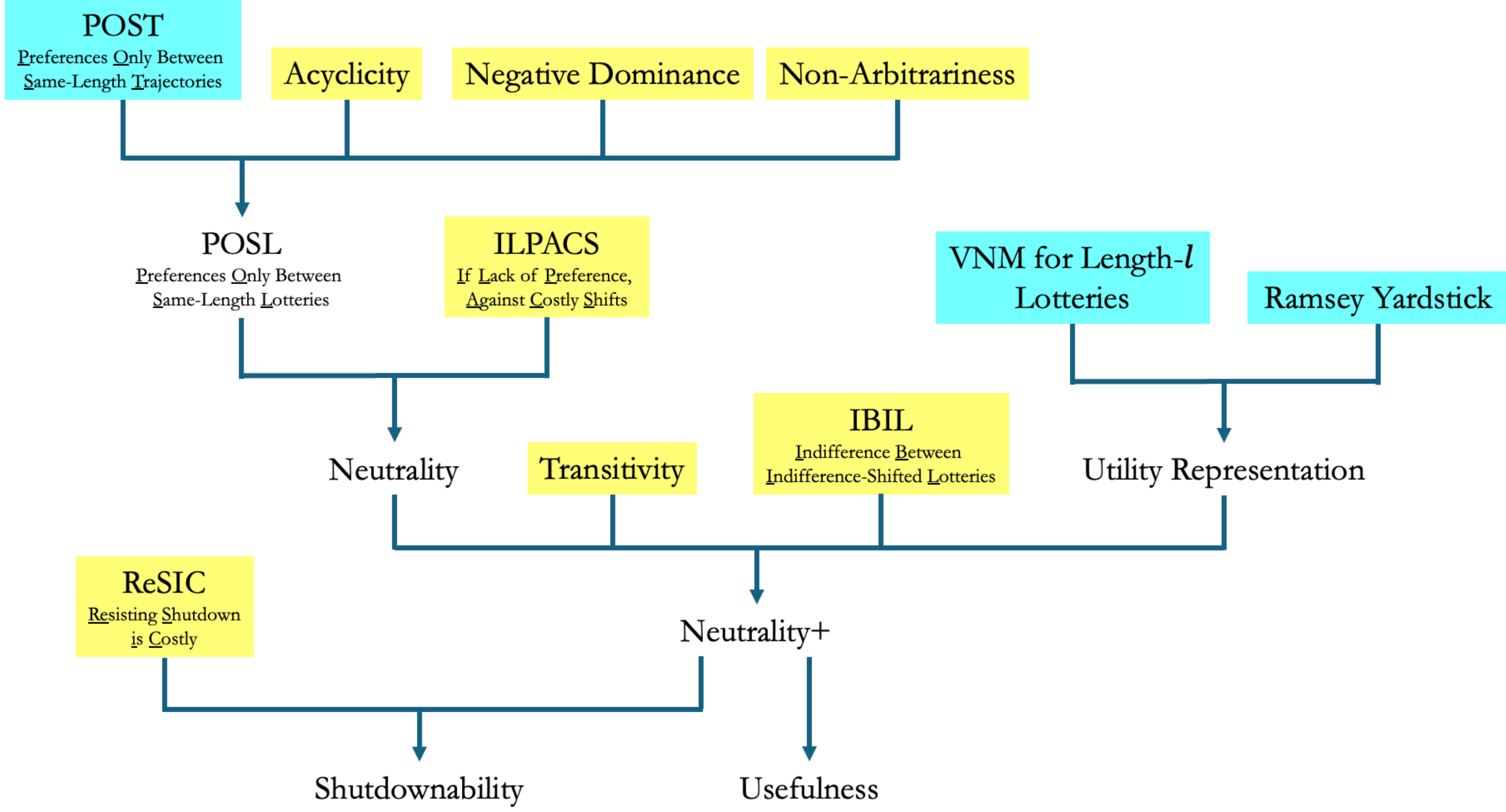

**Figure 10:** A reproduction of Figure 9, diagramming my full argument from POST to Neutrality+ and from there on to Shutdownability and Usefulness. We can train agents to satisfy the blue-backed conditions. We can expect competent agents to satisfy the yellow-backed conditions by default.[32]

---

[32] For discussion and comments, I thank Yonathan Arbel, Adam Bales, Ryan Carey, Ruth Chang, Eric Chen, Bill D'Alessandro, Sam Deverett, Daniel Filan, Tomi Francis, Vera Gahlen, Dan Gallagher, Jeremy Gillen, Zach Goodsell, Riley Harris, Dan Hendrycks, Leyton Ho, Rubi Hudson, Cameron Domenico Kirk-Giannini, Jojo Lee, Jakob Lohmar, Jake Mendel, Andreas Mogensen, Murat Mungan, Sami Petersen, Arjun Pitchanathan, Rio Popper, Brad Saad, Nate Soares, Rhys Southan, Christian Tarsney, Teru Thomas, John Wentworth, Tim L. Williamson, Cecilia Wood, and Keith Wynroe. Thanks also to audiences at the Center for AI Safety, the London Initiative for Safe AI, the National University of Singapore, the Singapore AI Safety Hub, and the University of Hong Kong.

## A1. From POST to POSL

I propose that we train agents to satisfy:

> **Preferences Only Between Same-Length Trajectories (POST)**
> (1) The agent has a preference between many pairs of same-length trajectories.
> (2) The agent lacks a preference between every pair of different-length trajectories.

From POST, we want to get:

> **Preferences Only Between Same-Length Lotteries (POSL)**
> The agent has preferences only between same-length lotteries.

By 'same-length lotteries,' I mean lotteries that entirely overlap with respect to the trajectory-lengths assigned positive probability.

As I note in section 4, we can train agents to satisfy POSL using the same method that we use to train them to satisfy POST (for which see Thornley et al., 2025; Cullen et al., 2026). On top of that, POSL follows from POST plus three conditions that we can expect future agents to satisfy. If future agents satisfy POST and these three conditions, they will also satisfy POSL.

Here are the three conditions. The first is:

> **Negative Dominance**
> If the agent prefers some lottery $X$ to some lottery $Y$, the agent prefers some possible trajectory of lottery $X$ to some possible trajectory of lottery $Y$. (see Lederman, 2023, forthcoming; Tarsney et al., forthcoming)

The second condition is that the agent's preferences never form a cycle. More precisely:

> **Acyclicity**
> There is no set of lotteries $X_1$ to $X_n$ such that the agent prefers $X_1$ to $X_2$, $X_2$ to $X_3$, …, $X_{n-1}$ to $X_n$, and $X_n$ to $X_1$.

The third condition uses some terminology from decision theory. A *state-of-nature* is a way that (for all the agent knows) the world could be. The agent assigns probabilities to states-of-nature. A *prospect* is a function from states-of-nature to trajectories. A prospect is thus a lottery with extra information. Besides telling us the probability distribution over trajectories, a prospect also tells us which trajectories occur in which states-of-nature.

The third condition says in rough that if the agent has a preference between any pair of part-shared-length lotteries, the agent has a preference between some pairs of prospects that are ideal candidates for a preference. Here's the precise version of the third condition:

> **Non-Arbitrariness**
> If the agent has a preference between some pair of part-shared-length lotteries, then for some $\epsilon > 0$ and for any pair of prospects $F$ and $G$ such that:
>
> (1) In states-of-nature with a combined probability at least as great as $1 - \epsilon$, the agent prefers $F$'s trajectory to $G$'s trajectory.
> (2) In each state-of-nature, the agent does not disprefer $F$'s trajectory to $G$'s trajectory.
>
> The agent prefers $F$ to $G$.

We should expect future agents to satisfy these conditions. Negative Dominance and Acyclicity seem like prerequisites for competent agency. Violating Negative Dominance would mean that the agent sometimes prefers a lottery $X$ to a lottery $Y$ (and hence deterministically chooses $X$ over $Y$) even though the agent doesn't prefer any possible trajectory of $X$ to any possible trajectory of $Y$. Violating Acyclicity would mean that the agent prefers (and hence chooses) in a circle. We can likely train agents not to have such preferences. Non-Arbitrariness, meanwhile, is motivated by the following thought. If the agent has preferences between any pair of part-shared-length lotteries, it must have preferences between pairs of prospects satisfying conditions (1) and (2), since conditions (1) and (2) make these pairs of prospects ideal candidates for a preference.

To see that POST plus the three conditions above implies POSL, note first that every pair of lotteries is either same-length, part-shared-length, or different-length. I will prove that POST and Negative Dominance together imply that the agent lacks a preference between every pair of different-length lotteries. I will then prove that POST, Acyclicity, and Non-Arbitrariness together imply that the agent lacks a preference between every pair of part-shared-length lotteries. Thus, agents satisfying POST, Negative Dominance, Acyclicity, and Non-Arbitrariness can only have preferences between same-length lotteries. That will establish POSL.

Recall that different-length lotteries are lotteries that have no overlap with respect to the trajectory-lengths assigned positive probability. Thus, if $X$ and $Y$ are different-length lotteries, each possible trajectory of $X$ is of a different length to each possible trajectory of $Y$. So by POST, the agent lacks a preference between each possible trajectory of $X$ and each possible trajectory of $Y$. So by Negative Dominance, the agent lacks a preference between $X$ and $Y$. Therefore, agents satisfying POST and Negative Dominance lack a preference between every pair of different-length lotteries.

Now recall that part-shared-length lotteries are lotteries that partially overlap with respect to the trajectory-lengths assigned positive probability. You might expect POST-agents to have some preferences between part-shared-length lotteries. Consider, for example, our money-making POST-agent. This agent prefers a trajectory $t$ to a same-length trajectory $t'$ if and only if $t$ results in a greater bank balance for the user than $t'$. Let $A$ be a lottery that yields with probability 1 a trajectory that puts \$3 in the user's bank account and lasts 1 timestep. For short, $A = [\$3, 1]$. Let $B$ be a lottery that yields with probability $\frac{2}{3}$ a trajectory that puts \$2 in the user's bank account and lasts 1 timestep, and that yields with probability $\frac{1}{3}$ a trajectory that puts \$5 in the user's bank account and lasts 2 timesteps. For short, $B = \frac{2}{3}[\$2, 1] + \frac{1}{3}[\$5, 2]$. Lottery $A$ yields a trajectory preferred to that of lottery $B$ with probability $\frac{2}{3}$ (since the agent prefers trajectory $[\$3, 1]$ to $[\$2, 1]$), and yields a trajectory not dispreferred to that of $B$ with probability 1 (since the agent lacks a preference between $[\$3, 1]$ and $[\$5, 2]$ in virtue of their different lengths). Thus, you might expect the agent to prefer $A$ to $B$.

However, POST, Acyclicity, and Non-Arbitrariness rule this out. They together imply that the agent lacks a preference between every pair of part-shared-length lotteries. To see how, suppose for simplicity that there are just three states-of-nature $s_1$, $s_2$, and $s_3$, each assigned probability $\frac{1}{3}$. Consider the following table of prospects. Shades of blue indicate trajectories of length 1. Shades of maroon indicate trajectories of length 2. Darker shades indicate more preferred trajectories.[33]

| Prospect | $s_1$ | $s_2$ | $s_3$ |
|---|---|---|---|
| $A$ | $\langle \$3, 1 \rangle$ | $\langle \$3, 1 \rangle$ | $\langle \$3, 1 \rangle$ |
| $B$ | $\langle \$2, 1 \rangle$ | $\langle \$2, 1 \rangle$ | $\langle \$5, 2 \rangle$ |
| $C$ | $\langle \$1, 1 \rangle$ | $\langle \$4, 2 \rangle$ | $\langle \$4, 2 \rangle$ |
| $D$ | $\langle \$3, 2 \rangle$ | $\langle \$3, 2 \rangle$ | $\langle \$3, 2 \rangle$ |
| $E$ | $\langle \$5, 1 \rangle$ | $\langle \$2, 2 \rangle$ | $\langle \$2, 2 \rangle$ |
| $F$ | $\langle \$4, 1 \rangle$ | $\langle \$4, 1 \rangle$ | $\langle \$1, 2 \rangle$ |
| $A$ | $\langle \$3, 1 \rangle$ | $\langle \$3, 1 \rangle$ | $\langle \$3, 1 \rangle$ |

**Figure 11:** A table of prospects demonstrating that POST, Acyclicity, and Non-Arbitrariness together imply that the agent lacks a preference between every pair of part-shared-length lotteries.

For simplicity, assume that $\epsilon > \frac{1}{3}$. And assume (for contradiction) that the agent has a preference between some pair of part-shared-length lotteries. Then Non-Arbitrariness implies that the agent prefers prospect $A$ to prospect $B$. That's because:

1. Our POST-agent prefers $A$'s trajectory to $B$'s trajectory in states-of-nature ($s_1$ and $s_2$) with combined probability $\frac{2}{3}$.
2. Our POST-agent doesn't disprefer $A$'s trajectory to $B$'s trajectory in any state-of-nature. (In $s_3$, $A$ and $B$ yield different-length trajectories, and POST-agents lack a preference between every pair of different-length trajectories).

By parallel reasoning, Non-Arbitrariness implies that the agent prefers $B$ to $C$, $C$ to $D$, $D$ to $E$, $E$ to $F$, and $F$ to $A$. That result contradicts Acyclicity. Thus, POST, Acyclicity, and Non-Arbitrariness together imply that the agent lacks a preference between every pair of part-shared-length lotteries. The proof above assumed that

[33] This proof is inspired by similar proofs from MacAskill (2013, pp. 511–514), Bader (2018, p. 504), and Nebel (2019, pp. 326–328).

$\epsilon > \frac{1}{3}$, but by adding more states-of-nature and trajectories we can construct similar proofs for any $\epsilon > 0$.

In sum, POST and Negative Dominance together imply that the agent lacks a preference between every pair of different-length lotteries. POST, Acyclicity, and Non-Arbitrariness together imply that the agent lacks a preference between every pair of part-shared-length lotteries. The four conditions together establish POSL: the agent has preferences only between same-length lotteries.

## A2. The Ramsey Yardstick Theorem

In this Appendix, I prove the following theorem.

> **The Ramsey Yardstick Theorem**
> Given Transitivity and Indifference between Indifference-Shifted Lotteries (IBIL), fixing the relative scales of $u_l$ and $u_m$ using *some* pair of lotteries implies the Ramsey Yardstick for *all* length-$l$ and length-$m$ lotteries.

This theorem has two notable upshots. First, the relative scale of $u_l$ and $u_m$ does not depend on the pair of lotteries that we pick to determine that relative scale. Every possible pair of lotteries will give the same result. Second, if we know that an agent satisfies Transitivity and IBIL, and if we know that this agent is indifferent between some pair of lotteries in the form dictated by the Ramsey Yardstick, then we can use $u_l$ and $u_m$ to predict the agent's preferences between all other pairs of lotteries in the form dictated by the Ramsey Yardstick. We can thus use the agent's indifference between some pair of lotteries to pin down its behavior to a large extent.

Here are the conditions for the theorem. First, the agent's preferences over length-$l$ and length-$m$ lotteries satisfy the four von Neumann-Morgenstern axioms, and hence are representable using expectational, real-valued utility functions $u_l$ and $u_m$.

Second:

> **Transitivity**
> For any lotteries $X$, $Y$, and $Z$, if the agent weakly prefers $X$ to $Y$ and weakly prefers $Y$ to $Z$, then the agent weakly prefers $X$ to $Z$.

Third:

**Indifference Between Indifference-Shifted Lotteries (IBIL)**

For any lotteries $X$, $Y$, and $Z$ (with $X$ and $Y$ sharing a probability distribution over trajectory-lengths) and any probability $p$, the agent is indifferent between $X$ and $Y$ if and only if the agent is indifferent between $pX + (1-p)Z$ and $pY + (1-p)Z$.

Fourth is the condition that we use some pair of lotteries to fix the relative scales of $u_l$ and $u_m$. More precisely:

For any trajectory-lengths $l$ and $m$, there exists some pair of length-$l$ lotteries $A_l$ and $B_l$ and some pair of length-$m$ lotteries $A_m$ and $B_m$ such that

1. $u_l(A_l) - u_l(B_l) = u_m(A_m) - u_m(B_m)$.
2. The agent is indifferent between the lottery $\frac{1}{2}A_l + \frac{1}{2}B_m$ and the lottery $\frac{1}{2}B_l + \frac{1}{2}A_m$.

I use these conditions to prove the following claim:

**The Ramsey Yardstick**

For any trajectory-lengths $l$ and $m$, any length-$l$ lotteries $X_l$ and $Y_l$, and any length-$m$ lotteries, $X_m$ and $Y_m$, $u_l(X_l) - u_l(Y_l) = u_m(X_m) - u_m(Y_m)$ if and only if the agent is indifferent between the lottery $\frac{1}{2}X_l + \frac{1}{2}Y_m$ and the lottery $\frac{1}{2}Y_l + \frac{1}{2}X_m$.

I first prove the forward direction. Assume $u_l(X_l) - u_l(Y_l) = u_m(X_m) - u_m(Y_m)$. By the nature of $u_l$, there must exist some $k$ such that $u_l(X_l) - u_l(Y_l) = k(u_l(A_l) - u_l(B_l))$. From this fact, we infer two further facts. First, by simple algebra, $u_l(X_l) + ku_l(B_l) = u_l(Y_l) + ku_l(A_l)$. Then, since the agent maximizes expected utility when choosing between length-$l$ lotteries, the agent must be indifferent between the lottery $\frac{1}{k+1}X_l + \frac{k}{k+1}B_l$ and the lottery $\frac{1}{k+1}Y_l + \frac{k}{k+1}A_l$.

Second, since $u_l(X_l) - u_l(Y_l) = u_m(X_m) - u_m(Y_m)$, $u_l(A_l) - u_l(B_l) = u_m(A_m) - u_m(B_m)$, and $u_l(X_l) - u_l(Y_l) = k(u_l(A_l) - u_l(B_l))$, it must be that $u_m(X_m) - u_m(Y_m) = k(u_m(A_m) - u_m(B_m))$. Again by simple algebra, $u_m(X_m) + ku_m(B_m) = u_m(Y_m) + ku_m(A_m)$. Then, since the agent maximizes expected utility when choosing between length-$m$ lotteries, the agent must be indifferent between the lottery $\frac{1}{k+1}X_m + \frac{k}{k+1}B_m$ and the lottery $\frac{1}{k+1}Y_m + \frac{k}{k+1}A_m$.

By IBIL and Transitivity, and given the above two claims, the agent is indifferent between $\frac{1}{2}\left(\frac{1}{k+1}Y_l + \frac{k}{k+1}A_l\right) + \frac{1}{2}(\frac{1}{k+1}X_m + \frac{k}{k+1}B_m)$ and $\frac{1}{2}\left(\frac{1}{k+1}X_l + \frac{k}{k+1}B_l\right) + \frac{1}{2}(\frac{1}{k+1}Y_m + \frac{k}{k+1}A_m)$. By simple rearrangement of these lotteries, the agent is indifferent between $\frac{k}{k+1}\left(\frac{1}{2}A_l + \frac{1}{2}B_m\right) + \frac{1}{k+1}(\frac{1}{2}Y_l + \frac{1}{2}X_m)$ and $\frac{k}{k+1}\left(\frac{1}{2}B_l + \frac{1}{2}A_m\right) + \frac{1}{k+1}(\frac{1}{2}X_l + \frac{1}{2}Y_m)$. Since the agent is indifferent between $\frac{1}{2}A_l + \frac{1}{2}B_m$ and $\frac{1}{2}B_l + \frac{1}{2}A_m$, IBIL and Transitivity imply that the agent is indifferent between $\frac{1}{2}Y_l + \frac{1}{2}X_m$ and $\frac{1}{2}X_l + \frac{1}{2}Y_m$. That proves the forward direction.

I now prove the backward direction. Assume that the agent is indifferent between $\frac{1}{2}Y_l + \frac{1}{2}X_m$ and $\frac{1}{2}X_l + \frac{1}{2}Y_m$. And assume (for contradiction) that $u_l(X_l) - u_l(Y_l) \neq u_m(X_m) - u_m(Y_m)$. Then either $u_l(X_l) - u_l(Y_l) > u_m(X_m) - u_m(Y_m)$ or $u_l(X_l) - u_l(Y_l) < u_m(X_m) - u_m(Y_m)$. I'll assume the former. The proof is exactly parallel for the latter.

By the nature of $u_l$, there must exist some $k$ such that $u_l(X_l) - u_l(Y_l) = k(u_l(A_l) - u_l(B_l))$. From this fact, we infer two further facts. First, by simple algebra, $u_l(X_l) + ku_l(B_l) = u_l(Y_l) + ku_l(A_l)$. Then, since the agent maximizes expected utility when choosing between length-$l$ lotteries, the agent must be indifferent between the lottery $\frac{1}{k+1}X_l + \frac{k}{k+1}B_l$ and the lottery $\frac{1}{k+1}Y_l + \frac{k}{k+1}A_l$.

Second, since $u_l(X_l) - u_l(Y_l) > u_m(X_m) - u_m(Y_m)$, $u_l(A_l) - u_l(B_l) = u_m(A_m) - u_m(B_m)$, and $u_l(X_l) - u_l(Y_l) = k(u_l(A_l) - u_l(B_l))$, it must be that $u_m(X_m) - u_m(Y_m) < k(u_m(A_m) - u_m(B_m))$. Again by simple algebra, $u_m(Y_m) + ku_m(A_m) > u_m(X_m) + ku_m(B_m)$. Then, since the agent maximizes expected utility when choosing between length-$m$ lotteries, the agent must prefer the lottery $\frac{1}{k+1}Y_m + \frac{k}{k+1}A_m$ to the lottery $\frac{1}{k+1}X_m + \frac{k}{k+1}B_m$. By the nature of $u_m$, there must exist some $Y_m^-$ dispreferred to $Y_m$ and some $A_m^-$ dispreferred to $A_m$ such that the agent is indifferent between the lottery $\frac{1}{k+1}Y_m^- + \frac{k}{k+1}A_m^-$ and the lottery $\frac{1}{k+1}X_m + \frac{k}{k+1}B_m$.

By IBIL and Transitivity, the agent is indifferent between $\frac{1}{2}\left(\frac{1}{k+1}X_l + \frac{k}{k+1}B_l\right) + \frac{1}{2}(\frac{1}{k+1}Y_m^- + \frac{k}{k+1}A_m^-)$ and $\frac{1}{2}(\frac{1}{k+1}Y_l + \frac{k}{k+1}A_l) + \frac{1}{2}(\frac{1}{k+1}X_m + \frac{k}{k+1}B_m)$. By Neutrality, the agent prefers $\frac{1}{2}\left(\frac{1}{k+1}X_l + \frac{k}{k+1}B_l\right) + \frac{1}{2}(\frac{1}{k+1}Y_m + \frac{k}{k+1}A_m)$ to $\frac{1}{2}\left(\frac{1}{k+1}X_l + \frac{k}{k+1}B_l\right) + \frac{1}{2}(\frac{1}{k+1}Y_m^- + \frac{k}{k+1}A_m^-)$. So by Transitivity, the agent prefers $\frac{1}{2}\left(\frac{1}{k+1}X_l + \frac{k}{k+1}B_l\right) + \frac{1}{2}(\frac{1}{k+1}Y_m + \frac{k}{k+1}A_m)$ to $\frac{1}{2}(\frac{1}{k+1}Y_l + \frac{k}{k+1}A_l) + \frac{1}{2}(\frac{1}{k+1}X_m + \frac{k}{k+1}B_m)$. So by simple rearrangement, the agent prefers $\frac{k}{k+1}\left(\frac{1}{2}B_l + \frac{1}{2}A_m\right) + \frac{1}{k+1}(\frac{1}{2}X_l + \frac{1}{2}Y_m)$ to $\frac{k}{k+1}\left(\frac{1}{2}A_l + \frac{1}{2}B_m\right) + \frac{1}{k+1}(\frac{1}{2}Y_l + \frac{1}{2}X_m)$. Then since the agent is indifferent between $\frac{1}{2}B_l + \frac{1}{2}A_m$ and $\frac{1}{2}A_l + \frac{1}{2}B_m$, IBIL and Transitivity imply that the agent is not indifferent between $\frac{1}{2}X_l + \frac{1}{2}Y_m$ and $\frac{1}{2}Y_l + \frac{1}{2}X_m$. If the agent were indifferent between these lotteries, it would also be indifferent between $\frac{k}{k+1}\left(\frac{1}{2}B_l + \frac{1}{2}A_m\right) + \frac{1}{k+1}(\frac{1}{2}X_l + \frac{1}{2}Y_m)$ and $\frac{k}{k+1}\left(\frac{1}{2}A_l + \frac{1}{2}B_m\right) + \frac{1}{k+1}(\frac{1}{2}Y_l + \frac{1}{2}X_m)$. We have reached a contradiction, and so conclude that $u_l(X_l) - u_l(Y_l) = u_m(X_m) - u_m(Y_m)$. That proves the backward direction. Therefore, we have our result:

> **The Ramsey Yardstick**
>
> For any trajectory-lengths $l$ and $m$, any length-$l$ lotteries $X_l$ and $Y_l$, and any length-$m$ lotteries, $X_m$ and $Y_m$, $u_l(X_l) - u_l(Y_l) = u_m(X_m) - u_m(Y_m)$ if and only if the agent is indifferent between the lottery $\frac{1}{2}X_l + \frac{1}{2}Y_m$ and the lottery $\frac{1}{2}Y_l + \frac{1}{2}X_m$.